\documentclass[mnsc,nonblindrev]{informs3-hide-journal}

 \OneAndAHalfSpacedXI

\usepackage{setspace}
\usepackage{natbib}
 \bibpunct[, ]{(}{)}{,}{a}{}{,}%
 \def\bibfont{\footnotesize}%
 \def\BIBand{and}%

\TheoremsNumberedThrough     

\EquationsNumberedThrough    

\MANUSCRIPTNO{} 

\usepackage[utf8]{inputenc}
\usepackage[margin=1in]{geometry}
\usepackage{amsmath,amssymb}
\usepackage{graphicx, enumerate, adjustbox}
\usepackage{pgfplots}
\pgfplotsset{compat=newest}
\DeclareUnicodeCharacter{2212}{−}
\usepgfplotslibrary{groupplots,dateplot}
\usepackage{tikz}
\usepackage{graphicx}
\usetikzlibrary{patterns,shapes.arrows}
\usepackage{algorithm}
\usepackage{multirow}
\usepackage{bm,bbm}
\usepackage{comment}
\usepackage{url} 
\usepackage{hyperref}
\hypersetup{
    colorlinks=true,
    citecolor=blue,
    linkcolor=blue,
    filecolor=magenta,      
    urlcolor=blue,
}
\allowdisplaybreaks

\definecolor{myred}{RGB}{178,51,51}
\newcommand{\aq}[1]{{\color{black}#1}}

\usepackage{breakurl}
\expandafter\def\expandafter\UrlBreaks\expandafter{\UrlBreaks
  \do\a\do\b\do\c\do\d\do\e\do\f\do\g\do\h\do\i\do\j%
  \do\k\do\l\do\m\do\n\do\o\do\p\do\q\do\r\do\s\do\t%
  \do\u\do\v\do\w\do\x\do\y\do\z\do\A\do\B\do\C\do\D%
  \do\E\do\F\do\G\do\H\do\I\do\J\do\K\do\L\do\M\do\N%
  \do\O\do\P\do\Q\do\R\do\S\do\T\do\U\do\V\do\W\do\X%
  \do\Y\do\Z\do\0\do1\do2\do3\do4\do5\do6\do7\do8\do9}

\usepackage{natbib}
 \bibpunct[, ]{(}{)}{,}{a}{}{,}%
 \def\bibfont{\small}%
 \def\BIBand{and}%

\usepackage{pgfplots}
\pgfplotsset{compat=1.18}
\usepgfplotslibrary{groupplots}
\usetikzlibrary{calc}
\definecolor{cm0}{RGB}{59,76,192}
\definecolor{cm1}{RGB}{123,159,249}
\definecolor{cm2}{RGB}{192,212,245}
\definecolor{cm3}{RGB}{242,203,183}
\definecolor{cm4}{RGB}{238,132,104}
\definecolor{cm5}{RGB}{180,4,38}

\newcommand{\mx}[1]{{\color{black}{#1}}}

\newcommand{\R}{{\mathbb{R}}}
\newcommand{\E}{{\mathbb{E}}}
\renewcommand{\P}{\mathbb{P}}
\newcommand{\I}{\mathbb{I}}

\newcommand{\Sscr}{{\mathcal S}}

\usepackage[capitalize]{cleveref}
\crefname{assm}{Assumption}{Assumptions}
\crefname{defn}{Definition}{Definitions}
\crefname{lem}{Lemma}{Lemmas}
\crefname{app}{Appendix}{Appendices}
\crefname{prop}{Proposition}{Propositions}
\crefname{problem}{Problem}{Problems}
\creflabelformat{problem}{#2(#1)#3}
\crefname{part}{part}{parts}
\crefname{examp}{Example}{Examples}
\crefname{thm}{Theorem}{Theorems}
\crefname{cor}{Corollary}{Corollaries}

\begin{document}


\RUNAUTHOR{Qi and Wang}

\RUNTITLE{The Scaling Paradox in Human-AI Collaboration}


\TITLE{The Scaling Paradox in Human–AI Collaboration}

\ARTICLEAUTHORS{%
\AUTHOR{Anyan Qi}
\AFF{Naveen Jindal School of Management, University of Texas at Dallas, USA, \EMAIL{axq140430@utdallas.edu}}
\AUTHOR{Mengxin Wang}
\AFF{Naveen Jindal School of Management, University of Texas at Dallas, USA, \EMAIL{mengxin.wang@utdallas.edu}}
} 

\ABSTRACT{%
\mx{The discovery of scaling laws has highlighted the extraordinary potential of AI systems with a striking empirical pattern: as AI systems scale, their capabilities tend to improve predictably. Yet, in real-world applications, AI rarely operates in isolation; instead, it often works alongside humans, raising the question of whether these gains persist in human–AI collaboration.} In this work, we develop an analytical model to examine \aq{when the empirical scaling benefits of AI translate into improved human–AI joint system performance.} We demonstrate that the performance of a human–AI system can scale positively as the AI scales up—provided that humans have an accurate perception of the AI’s capabilities. \mx{Human misperception, however, can fundamentally alter this relationship: i) when humans over-perceive the AI’s capabilities, a scaling paradox may arise, in which greater AI scale reduces overall system performance and amplifies firm-level profit losses, and (ii) when humans under-perceive the AI’s capabilities, performance still improves with scale but at a substantially slower rate. We further show that firms can actively manage these distortions through operational policies such as cost internalization and perception alignment, whose effectiveness depends on the economics of AI deployment and the direction of human misperception. }These findings suggest that organizations may benefit more from managing the human–AI interface than from simply investing in larger, more expensive AI systems. More broadly, our results suggest that AI scaling should be viewed not only as a technological challenge, but also as a behavioral and operational one, and caution against the view that larger AI systems will automatically lead to better operational outcomes. Whether AI scaling creates value ultimately depends on how increased AI capabilities shape human beliefs and collaborative efforts.
}%



\maketitle

%

\section{Introduction}
\label{sec:intro}

The recent progress of artificial intelligence (AI) has been driven by a remarkably simple empirical pattern: when AI models are made larger, trained on more data, and supported by more computation, their performance tends to improve in a predictable way. This phenomenon, commonly known as the \emph{Scaling Law} \citep{kaplan2020scaling,hestness2017deep}, has become a central principle in modern AI development. It suggests that better AI performance comes not only from isolated algorithmic breakthroughs, but can also emerge systematically from increasing scale. For firms, this pattern is deeply appealing. If performance improves reliably with scale, then investing in larger models, more compute, and more extensive deployment appear to offer a clear path toward productivity gains. Scaling laws therefore provide both a scientific explanation for the rapid progress of AI systems and a managerial rationale for continued investment in increasingly capable AI technologies.

\smallskip

Yet, the promise of scaling laws is often interpreted as if AI systems operate in isolation. In many benchmark settings, a model is evaluated on its own: it receives an input, produces an output, and is scored according to measures such as accuracy, loss, or task completion. This is typically where the empirical scaling effect is observed. As a result, much of the discussion around scaling focuses on overcoming physical constraints in data, computation, and resources to further improve model performance. In operational settings, however, AI rarely works alone. It is embedded in workflows that involve human decision makers. For example, a physician may use an AI diagnostic tool, but still decides whether to trust its recommendation. A lawyer may rely on a generative AI system to draft legal arguments, but remains responsible for verifying the citations. A customer-service agent may receive AI-generated suggestions, but chooses how much effort to spend reviewing and modifying them. A software engineer may use a coding assistant, but still allocates attention to debugging, testing, and integration. In these settings, the relevant unit of analysis is not the AI model alone, but the \emph{human-AI system}.

\smallskip

At first glance, a human-AI system should scale with the capability of its AI component: as AI becomes more capable, humans should be able to spend less time on each task and redirect the saved capacity toward completing more tasks. Recent evidence, however, suggests that this relationship is neither automatic nor always positive. In some settings, optimism surrounding generative AI has encouraged people to scale AI-enabled workflows aggressively. This pattern is especially visible in software development, where agentic coding tools have fueled expectations that software engineers may soon become much less necessary; for instance, Anthropic's CEO predicted that AI could soon write 90\% of code, and eventually nearly all code \citep{amodei2025code}. Similar optimism has appeared in customer service, where Klarna publicized that its AI chatbot could handle work equivalent to hundreds of customer-service agents \citep{doerer2025klarna}. In scientific research, there is similarly widespread optimism that AI systems could accelerate scientific discovery by helping researchers generate hypotheses, analyze large datasets, and identify patterns that humans might otherwise overlook \citep{nationalacademies2025ai}. Yet realized productivity effects have been more mixed. A randomized field experiment by METR found that experienced open-source developers expected AI tools to make them more than 20\% faster, and even after the experiment believed that AI had saved them time, but in fact completed tasks 19\% more slowly when AI tools were available \citep{becker2025metr}. Klarna later moved to bring human workers back into the loop, emphasizing that the speed of AI still needed to be complemented by human empathy and service quality \citep{doerer2025klarna}. \citet{naddaf2025ai} similarly reports concerns that LLM tools have contributed to an explosion of low-quality biomedical research papers.

\smallskip

Even when AI scaling does not generate outright negative effects, another operational problem frequently arises: firms may invest heavily in increasingly advanced AI systems, yet employees fail to adjust their behavior enough to realize the potential gains. A 2026 report found that despite record-level enterprise AI investment, 54\% of workers had bypassed AI tools and completed tasks manually in the previous month, while another 33\% had not used AI at all \citep{walkme2026adoption}. Another survey similarly found that 62\% of employees viewed AI as significantly overhyped, even though 86\% admitted that they were not using AI tools to their full potential \citep{goto2025pulse}. At the firm level, despite \$30-\$40 billion in enterprise GenAI investment, 95\% of organizations reported zero return \citep{challapally2025genai}. These examples illustrate that, even when AI becomes more capable on paper, the human-AI system may fail to capture the speed of scaling.

\smallskip

\emph{Why can AI scaling fail in collaborative production systems? What distinguishes an AI-only system from a human-AI system in ways that may break the scaling effect? Are these empirical patterns merely coincidental, or do they reflect a systematic mechanism? Finally, what operational interventions can preserve the benefits of scaling in a human-AI system?} Motivated by these questions, we develop an analytical model of human-AI collaboration to examine whether the empirical scaling benefits of AI necessarily translate into improved joint system performance. Specifically, we bring empirical scaling laws into a parsimonious model of human-AI collaboration. In our framework, an AI system with scale~$s$ and scaling factor~$\alpha$ works jointly with a human who has limited capacity and must decide how much effort to allocate to each project. The model captures the operational trade-off between improving the success probability of an individual project and increasing the total number of projects completed. We first show that the scaling law holds in a joint system when the human accurately perceives the AI's capability. AI scaling improves joint system performance because the human optimally reallocates effort toward greater throughput.

\smallskip

We then show that this conclusion can fail under a major behavioral mechanism: \emph{human misperception}. When the human \emph{over-perceives} AI capability, the human withdraws effort too aggressively, generating a \emph{scaling paradox}: increasing AI scale may reduce total system performance over a certain range of AI scale. When the human \emph{under-perceives} AI capability, scaling remains beneficial but the gains are substantially weakened because the human over-invests effort in each project. 
\mx{The scaling paradox can be carried over to the firm level. Because firms bear AI deployment costs, workers’ individually optimal effort choices may not maximize firm profit, even when they accurately assess AI capability. We show that human misperception can further distort this firm-worker misalignment. In particular, over-perception leads workers to reduce effort too aggressively, widening the gap between worker- and firm-optimal behavior and amplifying the negative effect of the scaling paradox on profit.} Under-perception, by contrast, may partially benefit the firm: by inducing the worker to exert more effort per project and undertake fewer projects, it can offset the underlying firm-worker misalignment and reduce the firm’s exposure to AI deployment costs. Our results highlight the asymmetry of misperception risk: over-perception can both reverse the beneficial effect of AI scaling and amplify the firm-worker misalignment, whereas under-perception generally preserves the scaling benefit and may even improve firm profit by narrowing the gap between worker and firm incentives.

\smallskip

We further examine how firms can mitigate these distortions through operational policy interventions. We study two organizational interventions that firms can use to mitigate scaling failure: \emph{cost internalization} and \emph{perception alignment}. Cost internalization shifts part of the AI deployment cost to workers, inducing them to exert more effort per project and discouraging inefficient broad AI use. We show that the optimal degree of cost internalization depends on the AI deployment cost and scale: Shifting the full deployment cost to workers can increase firm profit when AI is inexpensive or deployed at a small scale. As deployment becomes more costly or AI scale increases, however, firms should absorb a larger share of the cost to maintain worker adoption and preserve the value of AI use. Perception alignment instead corrects workers' beliefs about AI capability. This intervention is particularly valuable under over-perception, as alignment prevents workers from withdrawing effort too aggressively. When under-perception is present, however, alignment need not always increase firm profit, because workers' additional effort may partially offset the firm's underlying preference for greater human involvement. 

\smallskip
Taken together, our results suggest that AI scaling should be viewed not only as a technological challenge, but also as an operational and organizational one. More broadly, our results caution against the view that larger AI systems will automatically produce better social and organizational outcomes. Realizing the value of a deployed AI system requires firms to adapt human beliefs and incentive design to the system’s capability. As AI continues to advance, understanding the interaction between technological progress and human behavior will be increasingly important for ensuring that improvements in technologies translate into meaningful organizational performance.

\medskip

\subsection{Contributions}
In summary, our paper studies how AI scaling affects the performance of human-AI production systems when human effort is allocated endogenously and perceptions of AI capability may be imperfect. While larger AI models are often presumed to improve operational outcomes, this presumption overlooks the behavioral and organizational responses that accompany AI deployment. We develop an analytical framework in which AI scale shapes technical capability, human collaborators adjust effort based on perceived capability, and firms internalize the cost of deploying AI at scale. This framework allows us to characterize when scaling improves joint performance, when it fails because of distorted human responses, and how firm-level incentives interact with human-AI collaboration. Our analysis yields four main contributions.

\smallskip

    First, we introduce scaling laws into operations management modeling. The scaling law has primarily been studied as an empirical regularity in machine learning, linking model performance to compute, data, or model size. We bring this idea into an analytical operations framework by embedding AI scale into a capacity-allocation model of human-AI collaboration. This integrated framework allows us to study not only whether the AI model improves with scale, but also whether the entire production system improves when human effort responds endogenously. To our knowledge, this work is among the first in the operations management literature to apply the scaling-law logic as a primitive for studying AI-enabled operations. 

\smallskip
 
    Second, we identify human misperception, a novel behavioral mechanism, through which AI scaling may fail. Existing discussions of AI scaling often focus on technical bottlenecks, such as data scarcity, compute costs, or model limitations. We show that scaling can also fail because of human misperception. When humans overestimate AI capability, they reduce their own effort too aggressively, and the resulting under-investment can dominate the direct technical gains from scaling. This mechanism generates a non-monotone relationship between AI scale and joint system performance, providing an explanation for why larger or more capable AI systems may produce worse organizational outcomes in collaborative settings.

\smallskip

    Third, we distinguish the asymmetric operational effects of human over-perception and under-perception. Both forms of misperception reduce total reward output from the human-AI system relative to perfect collaboration, but they do so in fundamentally different ways. Over-perception can reverse the benefits of scaling over a certain region while under-perception preserves monotonic improvement at a slower rate. This asymmetry is more pronounced at the firm level. Over-perception widens the incentive misalignment between the firm and the worker, thereby amplifying the adverse effect of the scaling paradox on firm profit, whereas under-perception may mitigate this misalignment and sometimes improve firm profit. This asymmetry has direct managerial relevance: the costs of over- and under-perception are not symmetric, and the appropriate interventions therefore differ. 

\smallskip

    Fourth, we contribute to the design of operational policies that mitigate distortions caused by human misperception of AI capability. We identify two organizational interventions—cost internalization and perception alignment—and show how their effective implementation depends on the direction of human misperception and the economics of AI deployment.

\medskip

This paper proceeds as follows. \cref{sec:literature} reviews the related literature. \cref{sec:model} introduces the modeling preliminaries. \cref{sec:results-human} analyzes human-AI collaboration under the different misperception mechanisms. \cref{sec:firm} turns to the firm perspective, and \cref{sec:firm_strategy} discusses the firm's policy interventions to mitigate scaling failure. \cref{sec:conclusion} concludes.

\medskip

\section{Literature Review}
\label{sec:literature}

\smallskip

\emph{Foundational Scaling Laws and Their Applications in Operations}.
A growing computer science literature documents predictable relationships between model scale and performance. \citet{hestness2017deep} provide broad empirical evidence that generalization error for deep learning follows predictable power-law scaling patterns across domains. Focusing on neural language models, \citet{kaplan2020scaling} show that cross-entropy loss scales as a power law with model size, dataset size, and training compute, and use these relationships to characterize compute-efficient training. \citet{hoffmann2022training} investigate the optimal model size and number of tokens for training a transformer language model under a given compute budget. Complementing these empirical studies, \citet{bahri2024explaining} develop a theoretical framework that explains neural scaling laws through variance-limited and resolution-limited regimes for both dataset size and model size.

\smallskip

Recent work has extended scaling-law analysis to operations research (OR) and operations management (OM) applications.  \citet{huang2025orlm} extend the scaling-law literature to optimization modeling, showing that LLM's optimization-modeling performance exhibits predictable improvements as model size and training data scale. \citet{wang2025efficient} use fine-tuning scaling laws to allocate limited labeled human data together with LLM-generated surrogates for causal inference. Both studies use scaling laws to improve an AI-driven application. In contrast, our paper embeds a scaling-law primitive into an analytical framework of human-AI collaboration.

\medskip

\noindent \emph{From AI Capability to Realized Performance}.
A growing empirical and experimental literature examines how AI affects worker performance when deployed as decision support rather than as a stand-alone technology. This literature provides mixed evidence. On the positive side, AI assistance has been shown to improve productivity in customer service, professional writing, software development, and high-skilled knowledge work \citep{brynjolfsson2025generative,noy2023experimental,peng2023copilot,cui2025highskilled}. At the same time, recent studies show that AI assistance can also generate heterogeneous or even negative effects when workers misjudge the tool's capabilities or when tasks fall outside the AI's effective frontier \citep{dell2026navigating,becker2025metr}. \cite{merali2025scaling} provides experimental evidence that productivity gains from LLM scaling diminish in agentic workflows requiring multi-step interactions, compared to non-agentic analytical tasks. In customer service operations, \cite{wang2026agentic} show that while AI deployment improves chat efficiency, it substantially lowers customer ratings. In process automation, \cite{beer2026behavioral} conduct behavioral experiments and show that workers who collaborate with robots take longer to complete their tasks. In demand planning, \citet{brau2024collaborative} show that properly designed human-machine collaboration processes can substantially outperform both standalone machine learning systems and conventional judgmental adjustment, suggesting that realized performance depends not only on AI capability. Together, these studies suggest that improvements in AI capability do not automatically translate into proportional improvements in performance achieved. This observation raises a central question for our study: whether the performance gains implied by AI scaling laws necessarily carry over to human-AI systems.

\medskip

\noindent \emph{Human Behavioral Responses to Algorithms.}
A related stream examines how humans adopt and deviate from algorithmic recommendations in operational workflows. One side of this literature emphasizes algorithm aversion. In retail pricing, \citet{caro2023believing} show that managers do not always follow revenue-maximizing recommendations from a decision-support system due to users' lack of trust in the tool. In healthcare operations, \citet{hou2024physician} show that physicians' adoption of an AI assistant depends on both AI smartness and AI transparency. Related studies in warehouse operations, retail replenishment, and ridesharing further show that workers often exercise discretion or resist algorithmic prescriptions, reducing the realized value of algorithmic decision support \citep{sun2025predicting,kawaguchi2021workers,liu2026algorithm}.
The opposite behavioral friction is the automation bias, which happens when humans over-rely on algorithm assistance. \citet{parasuraman2010complacency} review evidence that humans may overuse imperfect automated decision aids, leading to both omission and commission errors. In clinical decision support, \citet{goddard2012automation} show that automation bias can cause users to accept automated recommendations too readily and fail to detect new errors introduced by the system. More recent evidence from AI-assisted diagnosis shows that incorrect AI outputs can mislead radiologists away from otherwise correct decisions \citep{bernstein2023can}, while AI assistance has heterogeneous effects across radiologists and can hurt some users \citep{yu2024heterogeneity}. These findings suggest that deviations from optimal reliance on algorithmic recommendations can substantially affect the realized value of automation. Our model builds on this behavioral insight by studying how misperceived AI scaling capability changes human effort allocation and thereby affects joint human-AI performance.

\medskip

\noindent \emph{Analytical Models of Human-AI Collaboration.}
Our paper is also related to analytical models of human-AI collaboration in operations management. \aq{Human–AI collaboration has attracted increasing attention in the operations management literature, with a rapidly growing body of research examining applications such as service systems \citep{yang2026better}, content creation \citep{hu2026ai}, and copyright \citep{yang2024generative}, among others.} \citet{boyaci2024humanmachine} model how machine input affects a human decision maker with limited cognitive capacity and show that machine advice can improve overall accuracy while changing the nature of decision errors and cognitive effort. \citet{fugener2025roles} develop a task-allocation framework in which AI can be used for automation, augmentation, or not at all, and show that the optimal role of AI depends on the type of complementarity between humans and AI. \citet{guan2025bestmachine} study incentive design in human-machine collaboration and show that a machine with the highest standalone accuracy may not maximize overall system performance because human responses and strategic interactions affect realized outcomes. \citet{lu2025augmenting} analyze AI-assisted demand forecasting and show that deploying a prediction machine may reduce managers' forecasting effort, creating a trade-off between information disclosure and managerial incentives. \cite{de2026your} analytically show that humans supervising machines may fail to learn whether the machine is superior because verification bias and lack of exploration create persistent uncertainty about machine quality. \citet{bastani2025human} study the contractual challenges of sustaining human oversight as AI becomes increasingly reliable, showing that stronger AI can paradoxically make human vigilance more difficult to incentivize. Our paper complements this literature by examining a unique angle: how AI scaling interacts with workers’ misperception of AI capability to shape endogenous effort allocation, thereby determining whether improvements in standalone AI capability translate into better human–AI system performance.

\smallskip

\section{Model Setup}
\label{sec:model}

We study a setting in which a firm deploys AI to support its human workers. Higher-scale AI systems provide greater capability but entail higher per-project deployment costs. Given the AI scale, the worker collaborates with the AI to complete projects. Specifically, in \cref{sec:model-workflow}, we introduce the human-AI workflow and characterize how the performance of the human-AI system depends on the AI scale. We then explain how human workers may misperceive the AI's capability in \cref{sec:model-misperception}. \mx{Finally, we formulate the worker's effort allocation problem when the worker accurately perceives or misperceives the AI's capability in \cref{sec:model-capacity-allocation}. Building on the human–AI system, we analyze the firm’s perspective and potential policy interventions in \cref{sec:model-firm}.}

\smallskip

\subsection{Human-AI Workflow and Scaling Law}\label{sec:model-workflow}
\smallskip

We begin by considering an AI agent and a human worker working together on a single project. Following the typical collaboration paradigm between human and AI, we model the collaboration as a stylized three-step sequential process: the human worker first initiates and sets up the project, the AI then attempts the project and produces initial output, which may not be correct, and the worker then reviews, validates, and when necessary, corrects the AI-generated output. Examples include software development, where engineers define the task and then test and review all AI-generated code; legal work, where lawyers specify the legal question and subsequently verify AI-generated drafts and citations; healthcare, where physicians provide the relevant clinical context and review all AI-generated recommendations; and customer service, where agents frame customer requests and review AI-drafted responses before sending them. In each setting, the worker must allocate review capacity to every AI-generated output because its correctness cannot be determined costlessly in advance. This stylized three-step structure captures the main feature of human-AI collaboration, abstracting from richer multi-step interactions that add sequential complexity without changing the core mechanism we study.

\smallskip

\mx{In the first step, the human worker sets up the project with a fixed setup time $t_0$. }The AI agent then attempts the project with success probability $1 - e^{-\alpha s}$, where $s \in \R^+$ denotes the AI agent's scale and $\alpha>0$ represents the AI scaling factor. This exponential functional form is motivated by the neural scaling laws literature, which shows that AI performance improves as a power law with model scale measured in size of compute or parameters \citep{kaplan2020scaling, hestness2017deep}. 
\mx{Specifically, let \(N>0\) denote the physical scale of an AI system along a given technological scaling path. Depending on the application, \(N\) may represent model size, computational resources, or an effective scale measure along which model size, data, and compute are jointly increased. The core feature of the standard empirical scaling law is that task-level failure measure follows an inverse power law in physical scale $N$, given by $q(N) = O(N^{-\alpha})$, where \(\alpha>0\) is a scaling exponent. For analytical convenience, we measure AI scale in logarithmic units $s = \ln(N)$. Focusing on the first-order component of task failure, we consider $N \geq 1$ and $s \geq 0$ and let $q(N) = N^{-\alpha} =e^{-\alpha\ln N} = e^{-\alpha s}$. Accordingly, the AI's probability of successfully completing a project is $1-e^{-\alpha s}$.}
Scale $s$ can thus be regarded as a measure of the total effective resources embodied in an AI system. For example, larger values of $s$ may correspond to more advanced model versions (e.g., moving from GPT-4 to GPT-5),  using high-reasoning models instead of fast-response models, or access to more capable AI services through premium plans.  

\smallskip

The scaling factor $\alpha$ captures how effectively AI scale translates into task performance. A larger value of $\alpha$ implies that increases in AI scale generate larger improvements in the probability of task success. Importantly, the scaling factor $\alpha$ is not only a characteristic of AI itself, but also a parameter that depends on how well the AI's capabilities align with the requirements of a particular task. Consequently, the same AI model may exhibit very different values of $\alpha$ across domains and applications. In practice, the effective value of $\alpha$ can be estimated from benchmark experiments that compare the performance of AI systems of different scales on a common set of tasks. In our work, we consider a human worker in the firm who employs the AI to complete one common set of projects, e.g., writing code.

\smallskip

Together, the two key parameters~$\alpha$ and $s$ determine the capability of the AI agent. Specifically, the AI's success probability, $1-e^{-\alpha s}$, is increasing in both parameters. A larger scale $s$ or a larger scaling factor $\alpha$ implies a greater likelihood that the AI can successfully complete the task without human intervention. When $s=0$, the success probability is zero, meaning that the AI provides no effective contribution to task completion. In contrast, as $\alpha s \rightarrow \infty$, the success probability approaches one, representing a fully capable AI agent that can complete the task almost certainly on its own. 

\smallskip

After the AI produces the initial output, the worker spends $t$ units of time reviewing, validating, and, when necessary, correcting the AI-generated output. Because the correctness of the AI output is not costlessly observable before review, the worker commits this review capacity to every project, regardless of whether the AI's initial output is ultimately correct. If the AI's output is incorrect, the worker successfully corrects it with probability $1-e^{-t}$. If the human worker also fails, the project is unsuccessful and the worker receives a payoff of zero; conversely, if the project succeeds—either through the AI agent alone or through the combined efforts of the AI agent and the human worker—the worker receives a payoff normalized to one. We assume that the success probabilities of the AI and the human are independent. Together, the project's success probability is 
\begin{equation*}
   p(t;\alpha,s) =  1 - e^{-\alpha s} + e^{-\alpha s}\left(1 - e^{- t}\right) = 1 - e^{-\alpha s - t}\,.
\end{equation*}

\smallskip

\subsection{Human's Misperception of AI's Capability}\label{sec:model-misperception}

\smallskip

In practice, AI's scale $s$ is often observable or can be reasonably inferred from public information such as model version, price, or compute level. We therefore treat $s$ as known to workers. However, the scaling factor $\alpha$ is much harder to understand and calibrate. The scaling factor is not a direct feature of the model itself, but a reduced-form parameter governing how effectively AI scale translates into task success. In practice, this mapping depends on the nature of the task, the extent to which the task matches the AI's strength, and the surrounding workflow in which the AI is deployed. As a result, even for a given model and scale, the effective value of $\alpha$ may vary substantially across contexts. 

\smallskip

Due to this lack of observability, \textit{individuals often infer $\alpha$ from their own mental models of AI or through limited interactions with AI -- which is seldom grounded in systematic evidence and can often be superstitious.} These subjective and often noisy judgments can lead workers to either overestimate or underestimate the AI's true scaling factor. We therefore let $\hat{\alpha}$ denote the worker's \emph{perceived} scaling factor, which may differ from the true $\alpha$.
Scenarios in which the human \emph{misperceives the AI’s capability}, $\hat{\alpha} \neq \alpha$, can be categorized into the following two types: \emph{over-perception} and \emph{under-perception}.

\smallskip

\begin{itemize}
\item \emph{Over-Perception}. Over-perception happens when humans \emph{overestimate} AI's capability: they infer that AI scale translates into task performance more effectively than it actually does. For instance, strong empirical scaling results on benchmark tasks may lead practitioners to believe that these gains generalize to a much broader set of untested tasks and real-world applications. Formally, this scenario corresponds to $\hat{\alpha} > \alpha$.

\smallskip
Empirical evidence of over-perception is abundant across professional settings. In a broad synthesis spanning aviation, medicine, and process control, \citet{parasuraman2010complacency} show that \textit{automation bias}---the tendency to over-rely on automated outputs and fail to detect errors one would otherwise catch---is pervasive even among trained experts and cannot be eliminated by standard interventions. A recent study by \citet{marcoccia2026ai} shows that unreliable AI advice can make people three times less accurate while leaving them twice as confident, suggesting that users may trust incorrect AI recommendations even as decision quality deteriorates. The legal profession provides a striking instance: in \textit{Mata v.\ Avianca} (S.D.N.Y.\ 2023), an attorney submitted AI-generated case citations to a federal court, involving fabricated quotations and internal citations---resulting in a court-imposed sanction \citep{mata2023avianca}. In each case, the human reduced their own effort and oversight in the belief that the AI was more capable than it truly was, directly related to the mechanism formalized in our model with $\hat \alpha>\alpha$.

\smallskip

\item \emph{Under-Perception}. Under-perception happens when humans \textit{underestimate} the AI's capability. This arises in contexts where AI is a relatively new entrant and practitioners remain skeptical. Formally, this scenario corresponds to $\hat{\alpha} < \alpha$.

\smallskip
The empirical literature on \textit{algorithm aversion} documents this counterpart systematically. \citet{dietvorst2015algorithm} provide experimental evidence that people are especially averse to algorithmic forecasters after seeing them perform, even when they see them outperform a human forecaster. The phenomenon is particularly pronounced in contexts resembling human-AI collaboration on knowledge tasks. \citet{castelo2019task} show---across four online lab studies with over 1,400 participants and two online field studies with over 56,000 participants---that consumers systematically discount algorithmic advice for tasks perceived as subjective or requiring uniquely human judgment, even when algorithms often outperform humans in those domains. Most consequentially for our model, \citet{filiz2023extent} document that algorithm aversion \textit{intensifies} as decision stakes rise: while 70.83\% of participants chose algorithmic guidance in less serious decision contexts, only 50.7\% did so in higher-gravity contexts. This ``tragedy of algorithm aversion'' means that human underestimation of AI is often most severe precisely where collaboration would yield the largest gains, precisely the mechanism formalized in our model with $\hat \alpha<\alpha$.
\end{itemize}

\medskip

\begin{remark}(Persistence of Misperception) One may wonder whether misperceptions disappear after repeated interactions with AI systems as workers update their beliefs based on feedback. In practice, however, learning the true capability of AI remains difficult, and misperceptions cannot be fully eliminated for various reasons:

\smallskip
First, unlike traditional technologies that are typically designed for a relatively narrow set of applications, modern foundation AI models are deployed across highly heterogeneous tasks. For example, language models such as GPT and DeepSeek can be used for document drafting, translation, customer support, software development, consulting, and even medical decision support. Because performance varies substantially across tasks and workflows, workers receive noisy and context-dependent feedback from their interactions with AI. Consequently, it becomes challenging for workers to infer the AI’s underlying capability parameters from their own interactions alone. In fact, rigorous AI evaluation has itself become a major open challenge in modern AI development \citep{maslej2025artificial}. At the same time, AI technologies evolve at an unprecedented pace. Newer models, reasoning systems, and specialized AI tools are introduced frequently, often within months rather than years \citep{maslej2025artificial}. The capabilities of AI systems are continually changing.  As a result, even when workers gradually update their beliefs based on experience, learning may not keep pace with technological progress.

\smallskip

Second, AI tools are increasingly used in high-stakes tasks, which further gives rise to the \emph{verification bias} mechanism identified by \cite{de2026your}. Decision-makers typically observe whether the AI was correct only when they follow its recommendation; they do not observe the counterfactual outcomes in cases where they override or ignore the AI. As a result, learning about the AI's capability can be biased and incomplete. This selective feedback may leave the decision-maker in a state of ``forever hesitation,'' in which she never obtains sufficient evidence to determine whether the machine is better. More fundamentally, this verification bias is closely related to the selective-labels problem \citep{lakkaraju2017selective}: in many human-AI systems, outcomes are observed only for cases in which humans make particular decisions, while the outcomes under alternative decisions remain missing. Such selective feedback is intrinsic to many human-AI systems and therefore cannot be fully eliminated through repeated use alone.

\smallskip

A third source of persistent bias may arise from psychological miscalibration. In highly personal domains such as medical diagnosis, patients often do not evaluate AI capability purely on objective performance. Instead, they may believe that their own case is unusual, context-specific, and difficult to reduce to standardized patterns, which leads them to perceive AI providers as less capable of accounting for their individual characteristics and circumstances than human providers \citep{longoni2019resistance}. This resistance is especially pronounced when tasks are perceived as subjective rather than objective, because users doubt that algorithmic systems can adequately incorporate judgment, empathy, and contextual nuance \citep{castelo2019task}. As a result, even repeated exposure to AI may not fully eliminate misperception: users may continue to discount AI's true capability, not because the AI performs poorly, but because they believe that only a human decision-maker can properly understand the uniqueness of their situation \citep{longoni2019resistance}.

\smallskip

Together, these reasons suggest that misperception is fundamental rather than transient and should therefore be explicitly analyzed. \hfill $\blacksquare$

\end{remark}

\medskip

\subsection{Human's Capacity Allocation}\label{sec:model-capacity-allocation} 

\smallskip
We next consider the human worker's capacity allocation decision. The human worker has a fixed working capacity $C$, representing the effective total working hours of the human worker. We first present the optimal benchmark capacity allocation problem where the human knows the true value of the AI scaling factor, i.e., $\hat \alpha = \alpha$. We then formulate the human's capacity allocation problem under misperception where $\hat \alpha \neq \alpha$.

\medskip

\noindent \emph{Human's Optimal Capacity Allocation Under $\hat \alpha = \alpha$. }Let $R(t; \alpha, s) := (1 - e^{-\alpha s - t}) \cdot \frac{C}{t + t_0}$ denote the total expected reward as a function of the human's capacity allocation $t$, under the true scaling factor $\alpha$ and AI's scale $s$. The human worker determines the optimal effort per project~$t$ based on AI's true capability $(\alpha, s)$ as  
\begin{equation}\label{eq:perfect-collaboration}
\max_{t \geq 0} ~R(t; \alpha, s) ~:=~ \left(1 - e^{-\alpha s - t}\right) \cdot \frac{C}{t + t_0}\,.
\end{equation}
Let $t^*(\alpha, s) := \argmax_{t \geq 0} R(t; \alpha, s)$ denote the optimal capacity allocation solution to \eqref{eq:perfect-collaboration} and let $R^*(\alpha, s) = R(t^*(\alpha, s); \alpha, s)$. $R^*(\alpha, s)$ represents the maximum reward achievable in this human-AI system. We refer to this setting as the \emph{perfect collaboration} scenario and use it as the optimal benchmark, which we discuss in more detail in \cref{sec:perfect-collaboration}. 

\medskip

\noindent \emph{Human's Capacity Allocation Under Misperception~$\hat \alpha \neq \alpha$.}
Given the known AI scale $s$ and the perceived AI scaling factor~$\hat{\alpha}$, the human worker determines the capacity allocation per project to maximize the \emph{perceived} total expected reward, which equals to the \emph{perceived} expected number of successful projects:
\begin{equation}\label{eq:human-optimal-effort}
    \max_{t \geq 0}~R(t; \hat{\alpha}, s) \,,
\end{equation}
where $R(t; \hat{\alpha}, s) =\left(1 - e^{-\hat{\alpha} s - t}\right) \cdot \frac{C}{t + t_0}\,.$
 Let $\hat{t}(\hat{\alpha}, s) := \argmax_{t \geq 0} R(t; \hat{\alpha}, s)$ denote the human's capacity allocation under misperception. The \textit{actual} expected reward realized by the joint system is then
\begin{equation}\label{eq:actual-reward}
    \hat{R}(\alpha, s) \;:= R(\hat{t}(\hat{\alpha}, s); \alpha, s) \;:=\; \bigl(1 - e^{-\alpha s - \hat{t}(\hat{\alpha}, s)}\bigr) \cdot \frac{C}{\hat{t}(\hat{\alpha}, s) + t_0}\,.
\end{equation}
Since $\hat{\alpha}\neq \alpha$, $\hat{t}(\hat{\alpha}, s)$ deviates from $t^*(\alpha, s)$. Consequently, the realized reward $\hat{R}(\alpha, s)$ may fall short of the optimal benchmark reward $R^*(\alpha, s)$. We characterize the solution to \eqref{eq:human-optimal-effort} and compare it with the benchmark in \eqref{eq:perfect-collaboration}  under over-perception (resp., under-perception) in \cref{sec:over-perception} (resp., \cref{sec:under-perception}). Specifically, we analyze how the realized expected reward varies with AI scale, examining whether scaling leads to monotonic improvements or whether misperception distorts this relationship.

\medskip

\begin{remark}(Project Heterogeneity)
Our baseline model assumes homogeneous projects with a common reward, setup time, and success probability. If projects are heterogeneous, workers would optimally allocate capacity differentially across projects, prioritizing those with higher rewards or lower effort requirements. Nevertheless, the central mechanism remains unchanged: misperceptions of AI capability continue to distort human effort allocation and thereby alter the realized gains from AI scaling. \hfill $\blacksquare$
\end{remark}

\medskip

\subsection{Firm's Perspective}  
\label{sec:model-firm}
\smallskip
We next proceed to analyze the human-AI system from the firm's perspective. The firm has an objective with a structure different from that of human workers: each project yields a net reward $r > 0$ if successful and zero otherwise, where the net reward is defined after subtracting the potential payoff to the human worker  (i.e., the unit reward received by the worker in the human-AI system). The deployment of AI at scale $s$ incurs a project cost of $c \cdot s$, where $c > 0$ is the cost per unit. This per-project cost structure captures a common feature of real-world AI deployment: inference is often priced on a usage basis, such as per input or output token, e.g., the API pricing of OpenAI (\url{https://developers.openai.com/api/docs/pricing}), Anthropic (\url{https://claude.com/pricing}), and Google Gemini (\url{https://ai.google.dev/gemini-api/docs/pricing}). In addition, each AI deployment (e.g., an API inference call) consumes computational resources that increase with the scale of the model.

\smallskip

We first consider a benchmark when the firm knows the value of $\alpha$ accurately and has full control over the worker's capacity allocation $t$. This corresponds to the firm’s best-case scenario, in which worker misperception is not a concern. The firm’s profit can then be written as
\begin{equation}\label{eq:firm_profit_full_control}
\Pi^{\mathrm{full}}(t; \alpha, s) \;:=\; \left(r\bigl(1 - e^{-\alpha s - t}\bigr) - c s\right) \cdot \frac{C}{t + t_0}\,.
\end{equation}

\smallskip

Given AI scale $s$, we let $t^{\mathrm{firm}}(\alpha, s) = \argmax_{t \geq 0} \Pi^{\mathrm{full}}(t; \alpha, s)$ denote the worker's capacity allocation that maximizes the firm's profit. Ideally, the firm would like the worker to work $t^{\mathrm{firm}}(\alpha, s)$ units of time on each project. However, there is an inherent firm-worker misalignment because the worker’s privately optimal allocation $t^*(\alpha,s)$ generally differs from $t^{\mathrm{firm}}(\alpha, s)$, as the firm bears the additional AI deployment cost whereas the worker does not. We analyze this misalignment in detail in \cref{sec:misalignment-perfect}. 

\smallskip

A larger gap between \(t^*(\alpha,s)\) and \(t^{\mathrm{firm}}(\alpha, s)\) implies a greater loss from the firm’s perspective. When worker misperception is present, this gap may change further, especially as the scale of the deployed AI system varies. 
We therefore study how misperception affects firm-worker misalignment and, in turn, the firm's profit. In particular, when the worker misperceives AI capability as \(\hat{\alpha}\), the resulting capacity allocation \(\hat{t}(\hat{\alpha}, s)\) may deviate further from \(t^{\mathrm{firm}}(\alpha, s)\), thereby amplifying the adverse effect of the scaling paradox on firm profit. We provide a detailed analysis in \cref{sec:misalignment-misperception}.

\smallskip

Lastly, in \cref{sec:firm_strategy}, we examine how firms can mitigate scaling failure through organizational interventions. We focus on two practical levers: \emph{cost internalization}, which shifts part of the cost of AI use to workers and thereby changes their effort-allocation incentives, and \emph{perception alignment}, which reduces the gap between workers' perceived and actual AI capability. 

\medskip

\section{Human-AI Collaboration: The Scaling Law Under Human (Mis)perception}
\label{sec:results-human}

\smallskip

In this section, we analyze the models in \cref{sec:model-capacity-allocation}, in which a human worker chooses the optimal capacity allocation for each project based on the perceived scaling factor $\hat{\alpha}$, and investigate the impact of human perceptions of AI capability on the scaling law. We begin with the benchmark case of perfect collaboration, where perception is accurate, in \cref{sec:perfect-collaboration}. We then examine the two forms of misperception—over-perception and under-perception—in \cref{sec:over-perception} and \cref{sec:under-perception}, respectively.

\smallskip

\subsection{Perfect Collaboration: \texorpdfstring{$\hat \alpha = \alpha$}{}}\label{sec:perfect-collaboration}

\smallskip

In this subsection, we analyze the benchmark setting \eqref{eq:perfect-collaboration} where the human worker is able to perfectly perceive the scaling factor of the AI. That is, $\hat \alpha = \alpha$. Recall that $t^*(\alpha, s)$ denotes the human worker's optimal capacity allocation decision under scale $s$ and scaling factor $\alpha$. The following proposition characterizes $t^*(\alpha, s)$.

\smallskip

\begin{proposition}[Optimal capacity allocation under perfect collaboration]\label[prop]{prop:optimal_t}
The total expected reward $R(t;\alpha, s)$ is quasi-concave with respect to $t \ge 0$. The optimal capacity allocation is
$$
t^*(\alpha, s) = \begin{cases}
-t_0 - 1 - W_{-1}\!\bigl(-e^{\alpha s - t_0 - 1}\bigr) & \text{if } \alpha s < \ln(1 + t_0) \\
0 & \text{if } \alpha s \geq \ln(1 + t_0)
\end{cases}\,,
$$
where $W_{-1}$ is the lower real branch of the Lambert $W$ function. When $\alpha s < \ln(1 + t_0)$, $t^*(\alpha, s)$ is the unique positive solution of 
\begin{equation}
\label{eq:optimal-t-foc}
    t + t_0 + 1 = e^{\alpha s + t}\,.
\end{equation}
\end{proposition}

\smallskip

The proposition identifies three important regimes, which we discuss below:

\smallskip

\begin{itemize}
    \item {\it Artificial General Intelligence (AGI) Regime}: This regime is characterized by $t^* = 0$, which occurs when $\alpha s \geq \ln(1+t_0)$. In this case, the AI is sufficiently capable that no human effort beyond the setup time is needed, and the throughput is limited only by $t_0$. This represents the ideal benchmark in which AGI-level performance has effectively been attained.

    \smallskip

    \item {\it Human-Only (HO) Regime}: The opposite extreme arises when $\alpha s = 0$. This occurs when the invested scale is negligible ($s=0$). 
    In this case, the capacity-allocation problem reduces to the human-only benchmark with success probability $1-e^{-t}$. The corresponding optimal capacity allocation, \(-t_0 - 1 - W_{-1}\!\bigl(-e^{-t_0-1}\bigr),\) provides an upper bound on human capacity allocation per project.

\smallskip

    \item {\it Human-in-the-Loop (HIL) Regime}: The intermediate case arises when $0 < \alpha s < \ln(1+t_0)$, for which the optimal human effort satisfies $t^*(\alpha,s) > 0$. In this regime, human effort is reduced relative to the human-only benchmark, but some human input remains necessary to achieve the optimal outcome. Given current obstacles to further scaling state-of-the-art AI systems---particularly the scarcity of high-quality human data---many real-world applications are best described by this regime. Thus, the economy lies between two limiting benchmarks: the human-only case with $\alpha s = 0$ and the AGI benchmark with $\alpha s \geq \ln(1+t_0)$. Our main focus is therefore on this intermediate region, in which human capacity allocation per project is strictly positive but smaller than in the human-only benchmark.

\end{itemize}

\smallskip

In the corollary below, we investigate the impact of AI parameters $\alpha$ and $s$ on the optimal decision, the project throughput, and the total reward. 

\smallskip

\begin{corollary}[Impact of $\alpha$ and $s$ under perfect collaboration]\label[cor]{cor:monotonicity}
Consider the optimal capacity allocation $t^*(\alpha, s)$ characterized in \cref{prop:optimal_t}. As AI scales,
\begin{enumerate}[(i)]
    \item the human effort is non-increasing, i.e., $t^*(\alpha, s)$ is non-increasing in both $s$ and $\alpha$;
    \item the project throughput is non-decreasing, i.e., the total number of projects $C\,/\,(t^*(\alpha,s)+t_0)$ is non-decreasing in both $s$ and $\alpha$;
    \item the total reward is non-decreasing, i.e., $R^*(\alpha, s)$ is non-decreasing in both $s$ and $\alpha$. In particular, the HIL system weakly outperforms the human-only benchmark, i.e., $R^*(\alpha, s) \geq R^*(\alpha, 0)$ for all $\alpha> 0, s \geq 0$.
\end{enumerate}
\end{corollary}

\smallskip

In summary, when humans have perfect knowledge of the AI's capability coefficient $\alpha$, the equilibrium outcome is consistent with our intuition: as the AI scales (indicated by an increased $\alpha s$), the human worker allocates less time per project, indicating an \emph{increased} total number of projects that they can work on, leading to an improved system performance relative to the system where humans work alone. Therefore, we conclude that \emph{the joint system's performance scales positively with the AI agent.} 

\smallskip

In practice, however, such an ideal scenario rarely happens. What is widely observed is that there are common misperceptions of AI's capability among humans, as we explained in Section~\ref{sec:model-misperception}. In particular, the human workers have limited knowledge of the ground-truth AI scaling factor~$\alpha$. \textit{Instead, individuals infer $\alpha$ from their own conceptual understanding of AI -- which is seldom grounded in systematic experience and is often superstitious.} Recall that we use $\hat{\alpha}$ to denote the human worker's \textit{perceived} AI scaling factor. In the following, we analyze the effect of misperception on the performance of the joint system. 

\smallskip

\subsection{Over-Perception: \texorpdfstring{$\hat{\alpha} > \alpha$}{}}\label{sec:over-perception}

\smallskip

In this subsection, we analyze the scenario where humans \textit{overestimate} the AI's capability---for instance, when AI performance on benchmarks leads practitioners to assume broader competence, or when media coverage inflates perceived abilities. Formally, this scenario corresponds to $\hat{\alpha} > \alpha$. Recall that, under over-perception, the human worker solves an optimization problem \eqref{eq:human-optimal-effort} on the perceived scaling factor $\hat{\alpha}$, yielding capacity allocation decision $\hat{t}(\hat{\alpha},s)$ and the actual realized reward $\hat{R}(\alpha, s)$.

\smallskip

In the next proposition, we compare the outcomes under the over-perception setting \eqref{eq:human-optimal-effort} and the perfect collaboration setting \eqref{eq:perfect-collaboration}.

\begin{proposition}[Comparison between over-perception and perfect collaboration]\label[prop]{prop:over_perception}
Under over-perception, i.e., $\hat{\alpha} > \alpha >0$, the following statements hold for all $s \geq 0$:
\begin{enumerate}[(i)]
    \item The human worker under-invests effort per project. That is, the human allocates weakly less time per project than the optimal under the true parameters: $\hat{t}(\hat{\alpha}, s) \leq t^*(\alpha, s)$.
    \item The actual reward falls below the perfect-collaboration benchmark: $\hat{R}(\alpha, s) \leq R^*(\alpha, s)$, with equality in the human-only regime ($s = 0$) and the AGI regime ($\alpha s \geq \ln(1+t_0)$). The inequality in the reward comparison is strict in the HIL regime where $0 <\alpha s <\ln(1+t_0)$. 
    \item \textbf{The scaling law fails:} There exists a threshold $\bar{\gamma} > 1$, which depends on $t_0$, such that whenever $\hat{\alpha}/\alpha > \bar{\gamma}$, the reward $\hat{R}(\alpha, s)$ is non-monotone in $s$: it initially increases at \(s=0\), but becomes non-monotone over the HIL interval \(s \in \bigl(0,\ln(1+t_0)/\alpha\bigr)\); in particular, it is strictly decreasing over some subintervals. Moreover, there exists \(0<\tilde{s}<\ln(1+t_0)/\alpha\) such that the human-AI system yields strictly less reward than the human-only benchmark, i.e., \(\hat{R}(\alpha,\tilde{s})<R^*(\alpha,0)\). 
\end{enumerate}
\end{proposition}

\smallskip
First of all, it is intuitive that when humans overestimate AI, they reduce their own effort per project faster than is warranted by the AI's true capability, as shown in \cref{prop:over_perception}.i. \emph{Prima facie}, since the human spends less time per project, they should have completed more projects and obtained a higher reward. However, because the human worker overestimates the capability of the AI, although the total number of projects increases, the success probability for each project is significantly reduced. In fact, the success probability is reduced so much that the actual reward drops to a level lower than that under the perfect collaboration setting, as shown in \cref{prop:over_perception}.ii.

\smallskip

Finally, a naive intuition would suggest that as the AI scales (by increasing $s$), the actual reward for the joint system of human and AI should continue to increase. Interestingly, we observe from \cref{prop:over_perception}.iii that the actual reward $\hat{R}(\alpha, s)$ can be non-monotone in $s$, indicating that the scaling law fails, particularly in the HIL interval. Moreover,  when the over-perception is sufficiently high, the reward in the human-AI system could be strictly worse than a human-only system when blindly scaling up the AI system. To understand this non-monotonicity, we note that there are two competing forces that drive the actual reward $\hat{R}(\alpha, s)$: the direct benefit of a larger-scale AI by increasing $\alpha s$ and the indirect cost of an increasingly insufficient human effort caused by reducing $\hat{t}(\hat{\alpha}, s)$
too aggressively. When the deployed AI scale increases without a corresponding adjustment in worker beliefs, realized performance can decline over part of the HIL region, since the reduction in human effort overpowers the direct gain through the larger-scale AI.

\smallskip

In the next corollary, we confirm the intuition that a higher degree of over-perception is detrimental to both the human effort per project and the actual reward of the joint system.

\smallskip

\begin{corollary}[Impact of the degree of over-perception]\label[cor]{cor:over_perception_comparative_static}
Fix \(\alpha>0\) and \(s\ge 0\). For each over-perceived capability \(\hat{\alpha}\in[\alpha,\infty)\), let
\(\hat{R}(\hat{\alpha};\alpha,s):=R(\hat{t}(\hat{\alpha},s);\alpha,s)\).
Then, for any \(\alpha\le \hat{\alpha}_1<\hat{\alpha}_2\),

\begin{enumerate}[(i)]
\item more severe over-perception further lowers human effort:
$t^*(\alpha,s)\ge \hat{t}(\hat{\alpha}_1,s)\ge \hat{t}(\hat{\alpha}_2,s);$

\item more severe over-perception further lowers actual reward: $R^*(\alpha,s)\ge \hat{R}(\hat{\alpha}_1;\alpha,s)\ge \hat{R}(\hat{\alpha}_2;\alpha,s).$
\end{enumerate}
\end{corollary}

\smallskip

\subsection{Under-Perception: \texorpdfstring{$\hat{\alpha} < \alpha$}{}}\label{sec:under-perception}

\smallskip

In this subsection, we analyze the other scenario where humans \textit{underestimate} the AI's capability. This often happens when AI is still new and practitioners remain skeptical. Formally, this scenario corresponds to $\hat{\alpha} < \alpha$. Similar to the analysis in the over-perception setting, the human again solves the optimization problem using $\hat{\alpha}$, yielding capacity allocation $\hat{t}(\hat{\alpha}, s)$, with actual reward $\hat{R}(\alpha, s)$. In the next proposition, we compare the equilibrium outcomes between the under-perception setting and the perfect collaboration setting. 

\smallskip

\begin{proposition}[Comparison between under-perception and perfect collaboration]\label[prop]{prop:under_perception}
Under under-perception, i.e., $\hat{\alpha} < \alpha$, the following statements hold for all $s \geq 0$:
\begin{enumerate}[(i)]
    \item The human worker over-invests effort per project. That is, the human allocates weakly more time per project than the optimal under the true parameters: $\hat{t}(\hat{\alpha}, s) \geq t^*(\alpha, s)$.
    \item \textbf{The scaling law holds with attenuated gains:} Scaling up the AI still benefits the joint system. That is, the actual expected reward $\hat{R}(\alpha, s)$ is non-decreasing in $s$. However, the actual reward falls below the perfect-collaboration benchmark: $\hat{R}(\alpha, s) \leq R^*(\alpha, s)$ for all $s \geq 0$. 
\end{enumerate}
\end{proposition}

\smallskip

When humans underestimate the AI's capability, they allocate excessive effort to each project relative to the true optimum, as shown in \cref{prop:under_perception}.i. This over-allocation reduces the number of projects that can be attempted and hence lowers overall productivity. Although scaling up the AI still improves the joint system's actual reward under under-perception, the gain is weaker than under perfect collaboration. In particular, the actual reward remains below the perfect-collaboration benchmark for all \(s \geq 0\), as shown in \cref{prop:under_perception}.ii. Thus, under-perception does not overturn the scaling law, but it leads to a less efficient utilization of larger-scale AI.

\smallskip

In the next corollary, we confirm the intuition that a higher degree of under-perception (i.e., a lower $\hat \alpha < \alpha$) inefficiently increases human effort per project and reduces the actual reward of the joint system.

\begin{corollary}[Impact of the degree of under-perception]\label[cor]{cor:under_perception_comparative_static}
Fix \(\alpha>0\) and \(s\ge 0\). For any \(0\le \hat{\alpha}_1<\hat{\alpha}_2\le \alpha\), 
\begin{enumerate}[(i)]
\item more severe under-perception further raises human effort: $\hat{t}(\hat{\alpha}_1,s)\ge \hat{t}(\hat{\alpha}_2,s)\ge t^*(\alpha,s);$

\item more severe under-perception further lowers actual reward: $\hat{R}(\hat{\alpha}_1;\alpha,s)\le \hat{R}(\hat{\alpha}_2;\alpha,s)\le R^*(\alpha,s).$
\end{enumerate}
\end{corollary}

\smallskip

Table~\ref{tab:summary} summarizes the qualitative effect of AI scaling across the three scenarios.
\begin{table}[ht]
\centering
\small
\caption{Effect of AI scaling ($s \uparrow$) on joint system performance under different human (mis)perceptions.}
\begin{tabular}{lccc}
\hline
Scenario & Time per project & Total attempted projects & Expected reward \\
\hline
Perfect collaboration ($\hat{\alpha} = \alpha$) & Decreasing & Increasing & Increasing \\
Over-perception ($\hat{\alpha} > \alpha$) & Steeply decreasing & Steeply increasing & \textbf{Non-monotone (can decrease)} \\
Under-perception ($\hat{\alpha} < \alpha$) & Slowly decreasing & Slowly increasing & Increasing (at slower rate) \\
\hline
\end{tabular}
\label{tab:summary}
\end{table}

\smallskip

\section{Amplified Impact of Scaling Failure on the Firm}
\label{sec:firm}

\smallskip

In the previous section, we discussed the effect of AI scaling on the joint system of the human worker and AI. In this section, we consider the firm's perspective. Recall from Section~\ref{sec:model-firm} that the firm’s optimal benchmark, defined in \eqref{eq:firm_profit_full_control}, assumes that the firm knows $\alpha$ exactly and has full control over the worker’s capacity allocation $t$. This benchmark is structurally different from that of the human worker. On the one hand, similar to the human worker, the firm benefits from successful projects and earns a net reward $r>0$ for each successful project. On the other hand, as the human worker employs AI, the firm needs to incur a per-project cost of $c\cdot s$ by deploying AI at scale~$s$, reflecting the operational fact that each AI deployment consumes computational resources proportional to the model's scale.

\smallskip

In what follows, we first consider the case where the AI scale~$s$ is exogenously given. Implicitly, the firm only considers $s$ such that $s < r/c$, since beyond the threshold the firm would gain nothing on each project. We start by analyzing the optimal effort allocation from the firm's perspective under perfect collaboration. Then, for each type of human misperception, we investigate how, as AI scales, human misperception contributes to the gap between the human effort allocation and the optimal effort allocation under perfect collaboration, as well as the firm's profit. 

\smallskip

\subsection{Firm-Worker Misalignment Under Perfect Collaboration: \texorpdfstring{$\hat \alpha=\alpha$}{}}
\label{sec:misalignment-perfect}

\smallskip

Due to the structurally different objectives of the firm and the human worker, we expect that the firm’s profit-maximizing effort allocation should differ from that when the human worker maximizes their own payoff. In this subsection, we analyze a hypothetical setting in which the firm can dictate the human effort allocation to maximize its profit in the perfect collaboration setting, i.e., $\hat \alpha=\alpha$. This optimal effort allocation from the firm's perspective, denoted by $t^{\mathrm{firm}}(\alpha, s)$, will serve as a benchmark for measuring the gap between the human worker's optimal effort allocation and the firm's profit-maximizing effort allocation. Formally, we define $t^{\mathrm{firm}}(\alpha, s)$ as the human effort allocation that would maximize the firm's profit for a given AI scale:
$$
t^{\mathrm{firm}}(\alpha, s) = \argmax_{t \geq 0} \; \left\{r\bigl(1 - e^{-\alpha s - t}\bigr) - cs\right\} \cdot \frac{C}{t + t_0}\,,
$$
which is characterized in \cref{lem:misalignment}.i below.

\smallskip

\begin{lemma}[Firm's profit-maximizing effort allocation under perfect collaboration]\label[lem]{lem:misalignment}
For any $s \in [0, \frac{r}{c})$:
\begin{enumerate}[(i)]
    \item The firm's profit is quasi-concave in $t$, and its unique maximizer is
        \[
        t^{\mathrm{firm}}(\alpha,s)=
        \left\{
        \begin{array}{ll}
       t \ge 0 \text{ such that }
        e^{-\alpha s - t}(t+t_0+1)=1-\dfrac{cs}{r},
        & \text{if } \alpha s < \ln\!\left(\dfrac{t_0+1}{1-\frac{cs}{r}}\right), \\[1.2em]
        0,
        & \text{if } \alpha s \ge \ln\!\left(\dfrac{t_0+1}{1-\frac{cs}{r}}\right).
        \end{array}
        \right.
        \]
    \item The firm prefers more human effort per project than the human voluntarily provides, i.e.,
    $t^{\mathrm{firm}}(\alpha, s) \geq  t^*(\alpha, s)$. The inequality is strict in the HIL regime. 
\end{enumerate}
\end{lemma}

\smallskip
\cref{lem:misalignment}.ii characterizes the consequence in the effort allocation due to the misaligned incentives between the firm and the human worker: the firm prefers the human worker to spend more effort on each project. The misalignment arises because each additional project the human undertakes incurs the AI cost $cs$, a cost the human ignores when maximizing the expected number of successes. The firm would prefer the human to work more on each project---increasing per-project success probability and reducing the number of costly AI deployments---rather than spreading capacity thin across many projects. This misalignment is strict in the HIL regime.

\smallskip

\mx{We next observe that the firm's profit-maximizing effort allocation $t^{\mathrm{firm}}(\alpha, s)$ in \cref{lem:misalignment} is positive only when scale $s$ lies in a certain region. For convenience, let us denote the two HIL regions---the regions of AI scales at which strictly positive human effort is required---for the firm and the human worker, respectively, by 
\begin{align*}
&\Sscr^{\mathrm{HIL, firm}} := \left\{s: 0 < s < \frac{\ln\left\{(t_0 + 1)/(1 - cs/r)\right\}}{\alpha}\right\};~
\Sscr^{\mathrm{HIL, human}} := \left\{s: 0 < s < \frac{\ln(t_0 + 1)}{\alpha}\right\}\,.
\end{align*}
It is immediate that unless the AI is free ($c=0$), we have that  $\Sscr^{\mathrm{HIL, human}} \subset \Sscr^{\mathrm{HIL, firm}}$, implying that the HIL region for the firm is strictly larger than that for the human worker. Intuitively, because the firm internalizes the AI deployment cost, it continues to value positive human effort even in regimes where the human worker would have already chosen to stop working. }

\smallskip

With the observation that the misalignment between the firm and the human worker is structural and persists even under perfect human-AI collaboration, we next analyze how misperceptions of AI capabilities further distort this inherent firm-worker misalignment.

\smallskip

\subsection{Firm-Worker Misalignment Under Misperception}
\label{sec:misalignment-misperception}

\smallskip

In this subsection, we examine how human workers' misperception of AI capability interacts with the firm's objective, starting from the \emph{over-perception} case where $\hat \alpha >\alpha$. Recall that even under correct beliefs, a structural firm-worker misalignment arises because the worker does not internalize the AI deployment cost. We show that over-perception systematically exacerbates this misalignment. In particular, not only does over-perception widen the gap between the firm's preferred effort under perfect collaboration and the human's chosen effort at a given scale, but this gap further increases as the AI system is scaled up.

\smallskip

For any $\hat{\alpha} \ge 0$ and $s \in [0, \frac{r}{c})$, define the misalignment gap
\[
\Delta(\hat{\alpha}, s) := \left|t^{\mathrm{firm}}(\alpha,s) - \hat{t}(\hat{\alpha},s)\right|\,.
\]

\begin{proposition}[Firm-worker misalignment under over-perception]\label[prop]{prop:misalignment_overperception}
For any $s \in [0,\frac{r}{c})$ and $\hat{\alpha} \ge \alpha$, the following statements hold:
\begin{enumerate}[(i)]
\item Firm-worker misalignment increases with the over-perception level, i.e., 
$\Delta(\hat{\alpha}, s)$ is weakly increasing w.r.t. $\hat{\alpha}$.
\item \mx{Firm-worker misalignment increases with AI scale $s \in I \cap \Sscr^{\mathrm{HIL, firm}}$, i.e., $\Delta(\hat{\alpha}, s)$ is weakly increasing w.r.t. $s \in I \cap \Sscr^{\mathrm{HIL, firm}}$, where $I = \{s\,:\,0 \le s<\frac{r}{c} ~\text{and}~ \frac{c}{r}(1+\alpha s)\ge \alpha\}$.}
\item The sensitivity of the firm-worker misalignment with respect to scale rises with the over-perception level, i.e., 
$\frac{\partial \Delta(\hat{\alpha},s)}{\partial s}$ \mx{is strictly increasing in $\hat{\alpha}$ in the region $s < \ln(t_0+1)/\hat{\alpha}$.}
\end{enumerate}
\end{proposition}

\smallskip

\cref{prop:misalignment_overperception}.i shows that, holding AI scale fixed, a higher level of over-perception weakly enlarges the misalignment gap \(\Delta(\hat{\alpha},s)\). Economically, the more human workers overestimate the AI’s capability, the more they reduce their own effort below the level preferred by the firm, thereby worsening the distortion that already exists even under perfect collaboration. \cref{prop:misalignment_overperception}.ii further shows that this misalignment also tends to grow with AI scale when certain conditions hold. The first condition, \(0\le s<r/c\), is the basic feasibility condition already imposed in the model: the per-project AI cost cannot exceed the reward. The second condition, $s \in \Sscr^{\mathrm{HIL, firm}}$, means that the AI scale is still in the firm’s HIL region, so from the firm’s perspective positive human effort remains valuable. The third condition, \(\frac{c}{r}(1+\alpha s)\ge \alpha\), is the most substantive one. This condition is more likely to hold either when \(s\) is relatively large or when the unit-cost-to-reward ratio \(c/r\) is relatively high. In other words, when AI is already deployed at a high scale, or when AI deployment is relatively costly, further scaling is more likely to enlarge the misalignment gap. Intuitively, in such regions the firm has stronger incentives to preserve human effort because each additional AI-assisted project is more expensive, whereas an over-perceiving worker cuts effort more aggressively as scale rises. \cref{prop:misalignment_overperception}.iii goes one step further by showing that over-perception not only increases the level of misalignment, but also amplifies how sensitive misalignment is to AI scale. That is, as \(\hat{\alpha}\) rises, the slope \(\partial \Delta(\hat{\alpha},s)/\partial s\) \mx{becomes strictly larger as long as the AI scale remains within the worker's \emph{perceived} HIL region. Once the AI scale exceeds this perceived HIL threshold, the worker exerts no effort, so $\hat{\alpha}$ no longer affects the misalignment. }

\smallskip

\begin{proposition}[Scaling law failure is amplified in the firm's profit]
\label{prop:profit_scaling_failure}
For $\hat{\alpha} \ge \alpha$, the following holds in the worker's perceived HIL region \(\hat{\alpha} s<\ln(1+t_0)\):
\(\Pi'(s;\alpha,\hat{\alpha})<
r\,\partial_s \hat{R}(\alpha,s)
\). As a result, whenever reward decreases with scale, i.e., $\partial_s\hat{R}(\alpha,s)<0$, profit decreases even more sharply, i.e., $\Pi'(s;\alpha,\hat{\alpha})<r\,\partial_s \hat{R}(\alpha,s) <0$.
\end{proposition}

\smallskip

Suppose the condition in Proposition~\ref{prop:over_perception}.iii holds, so that the scaling law fails, i.e., $\hat{R}(\alpha,s)$ decreases in $s$. Then $\Pi(s;\alpha,\hat{\alpha})$ is more distorted by scaling than $\hat{R}(\alpha,s)$ by Proposition~\ref{prop:profit_scaling_failure}. 
Hence, once scaling begins to reduce the realized outcome of the human-AI system, it reduces the firm's profit even more, since the firm additionally bears the direct deployment cost $cs$ on an increasing number of projects.

\smallskip

Taken together, Propositions~\ref{prop:misalignment_overperception} and \ref{prop:profit_scaling_failure} establish a compounding effect of over-perception. On the one hand, over-perception can widen the firm-worker misalignment, moving the firm further away from the optimal outcome under perfect collaboration characterized in \cref{lem:misalignment}.i. On the other hand, in the imperfect setting where workers misperceive AI capability, over-perception amplifies the consequences of scaling law failure discussed in \cref{prop:over_perception}.

\smallskip

This amplification mechanism has important implications for firm performance. Importantly, this mechanism runs counter to many real-world organizational practices. In practice, firms themselves often actively encourage or even induce over-perception of AI capabilities. For example, recent reporting indicates that some organizations evaluate employee performance using usage-based metrics for generative AI tools—such as the volume of tokens processed or the frequency of AI-assisted workflows—thereby incentivizing employees to rely more heavily on AI regardless of its true marginal productivity \citep{nytimes_tokenmaxxing_2026}. Meanwhile, reporting highlights that while employers emphasize AI-driven productivity gains, many workers experience increased errors, lower-quality outputs, and reduced oversight as reliance on AI expands \citep{guardian_ai_productivity_errors_2026}.

\smallskip

Such incentives can systematically reinforce over-perception: employees are not only prone to overestimate AI capabilities, but are also rewarded for acting on those beliefs. In light of our model, this creates a particularly concerning amplification mechanism:  Because over-perception already leads to inefficient reductions in human effort, organizational pressure to increase AI reliance can further magnify this distortion. As a result, the scaling failure identified in Proposition~\ref{prop:profit_scaling_failure} is likely to be more pronounced in practice: firms may push toward larger-scale deployment precisely in regimes where marginal returns are already declining, thereby accelerating the deterioration in realized performance and, even more sharply, in profit.

\smallskip

We next analyze the firm-worker misalignment in the \emph{under-perception} case, i.e., $\hat \alpha < \alpha$. By \cref{lem:misalignment}, we observe that under perfect collaboration, the firm-preferred effort is higher than the worker-chosen effort. By \cref{prop:under_perception}, we note that the human worker exerts more effort under under-perception than under perfect collaboration. Thus, one may expect that some degree of under-perception might help mitigate the firm-worker misalignment. In the next proposition, we show that under-perception indeed may alleviate but does not eliminate the misalignment in most cases.

\smallskip

\begin{proposition}[Firm-worker misalignment under under-perception]\label{prop:misalignment_underperception}
For any $s \in [0,\frac{r}{c}) \cap \Sscr^{\mathrm{HIL, firm}}$,
define
\[
\hat{\alpha}^{\dagger}
:=
\alpha+\frac{1}{s}\ln\!\left(1-\frac{cs}{r}\right)
<\alpha\,.
\]
Then the following statements hold:
\begin{enumerate}
    \item[(i)] $\Delta(\hat{\alpha}, s)$ is strictly decreasing w.r.t. $\hat{\alpha}$ when $\hat{\alpha} \in [0, \hat{\alpha}^{\dagger}]$, with $\Delta(\hat{\alpha}^{\dagger}, s) = 0$.
    \item[(ii)] $\Delta(\hat{\alpha}, s)$ is strictly increasing w.r.t. $\hat{\alpha}$ when $\hat{\alpha} \in [\hat{\alpha}^{\dagger}, \alpha]  $.
\end{enumerate}

\end{proposition}

\smallskip

The threshold \(\hat{\alpha}^{\dagger}\) characterizes when the human’s perceived AI capability exactly aligns their effort choice with the firm’s preferred level.
When \(\hat{\alpha}<\alpha\), the human underestimates the AI’s capability and therefore responds by working more. This increase in effort moves behavior in the direction preferred by the firm, partially mitigating the misalignment. However, this correction is generally imperfect: unless \(\hat{\alpha}=\hat{\alpha}^{\dagger}\), the human’s effort still deviates from the firm’s optimum.
If \(\hat{\alpha}>\hat{\alpha}^{\dagger}\), the human still under-invests in effort relative to the firm’s preferred level, despite under-perception relative to the true \(\alpha\). If \(\hat{\alpha}<\hat{\alpha}^{\dagger}\), the human overcompensates and exerts too much effort. Only at \(\hat{\alpha}=\hat{\alpha}^{\dagger}\) does the perceived trade-off exactly match the firm’s objective, eliminating the misalignment.

\smallskip

Compared to over-perception, under-perception is less risky for the firm. When \(\hat{\alpha}>\alpha\), the human systematically under-invests in effort, which unambiguously widens the firm-worker misalignment and can significantly reduce project success probabilities. By contrast, when \(\hat{\alpha}<\alpha\), the human responds by exerting more effort per project, which might move effort in the direction preferred by the firm. However, the overall consequence of under-perception remains ambiguous. 

 \smallskip

\section{Firm's Policy Interventions to Mitigate Scaling Failure}\label{sec:firm_strategy}
\smallskip
This section examines how firms can mitigate scaling failure through organizational interventions. We focus on two practical levers: cost internalization and perception alignment.

\smallskip

\subsection{Cost Internalization}
\smallskip

One practical policy is \emph{cost internalization} (CI), under which the firm shifts part of the cost of AI use to the worker. In practice, firms differ widely in how these costs are allocated. At one end, some firms explicitly subsidize enterprise AI access---for example, PwC announced a rollout of ChatGPT Enterprise to more than 100{,}000 employees in its U.S. and U.K. businesses, and Microsoft reportedly launched Copilot internally to encourage employee adoption \citep{reuters2024pwcopenai,reuters2024microsoftcopilot}. 
At the other end, firms may require workers to bear the cost of accessing employer-mandated AI tools. In one reported case, a product startup instructed its developers to purchase \$20-per-month Cursor subscriptions for assigned unit-testing tasks and later conditioned reimbursement on detailed documentation of tool usage and output. As a result, some employees ultimately paid the subscription costs out of pocket \citep{ndtv2025cursor}. This example is consistent with broader survey evidence: Deloitte reports that 31\% of U.K. employees who use generative AI at work personally pay for access \citep{deloitte2024genaipay}. CI captures this variation in cost allocation and allows us to study how shifting more of the deployment cost to workers changes effort allocation and firm outcomes.

\smallskip

Formally, let $\rho \in [0,1]$ denote the share of the unit AI deployment cost borne by the worker. Under CI, the worker chooses effort $t$ per project under the perceived scaling factor $\hat{\alpha}$ while internalizing a fraction $\rho$ of the AI cost:
\[
\hat t(\hat{\alpha}, s, \rho)
=
\arg\max_{t\ge 0}
\left[
(1-e^{-\hat{\alpha}s-t})-\rho c s
\right]\frac{C}{t+t_0}\,.
\]

Given $\hat t(\hat{\alpha}, s, \rho)$, the worker anticipates a payoff of
\[
\hat R^{\rho}
=
\left[
(1-e^{-\hat{\alpha}s-\hat t(\hat{\alpha}, s, \rho)})-\rho c s
\right]
\frac{C}{\hat t(\hat{\alpha}, s, \rho)+t_0}\,.
\]

 \mx{Note that, for expositional simplicity, we treat $\hat t(\hat{\alpha},s,\rho)$ as an extended-valued optimizer. When $\rho c s$ is sufficiently large, the objective may have no finite maximizer and instead approach its supremum as $\hat t(\hat{\alpha},s,\rho) \to\infty$. In this case, we set $\hat t(\hat{\alpha},s,\rho)=\infty$, implying an anticipated payoff of $\hat R^\rho=0$. }
The worker then compares this anticipated payoff with the payoff from a human-only system:
\[
\hat t^{\sf HO}
=
\arg\max_{t\ge 0}
\left[(1-e^{-t})\right]\frac{C}{t+t_0}\,,
\qquad
R^{\sf HO}
=
\left[(1-e^{-\hat t^{\sf HO}})\right]\frac{C}{\hat t^{\sf HO}+t_0}\,.
\]

The worker adopts the AI tool only when anticipating a higher payoff than the human-only system, i.e., $\hat R^{\rho} > R^{\sf HO}$. Accordingly, the worker's realized payoff is
\[
\left[(1-e^{-\alpha s-\hat t(\hat{\alpha}, s, \rho)})-\rho c s\right]
\frac{C}{\hat t(\hat{\alpha}, s, \rho)+t_0}
I\{\hat R^{\rho} > R^{\sf HO}\}
+
(1-e^{-\hat t^{\sf HO}})
\frac{C}{\hat t^{\sf HO}+t_0}
I\{\hat R^{\rho} \le R^{\sf HO}\}\,,
\]
while the firm's realized profit is
\[
\left[
r\bigl(1-e^{-\alpha s-\hat t(\hat{\alpha}, s, \rho)}\bigr)-(1-\rho)c s
\right]
\frac{C}{\hat t(\hat{\alpha}, s, \rho)+t_0}
I\{\hat R^{\rho} > R^{\sf HO}\}
+
r(1-e^{-\hat t^{\sf HO}})
\frac{C}{\hat t^{\sf HO}+t_0}
I\{\hat R^{\rho} \le R^{\sf HO}\}\,.
\]

\begin{figure}[ht]
  \centering 
  \caption{Cost Internalization Under Low AI Deployment Cost}
  \label{fig:cost_interlization_low_cost}
  \resizebox{1\linewidth}{!}{%
\input{fig/cost_internalization_r1_c001_ah1.tex}
}
\end{figure}

\begin{figure}[ht]
  \centering
  \caption{Cost Internalization Under Moderate AI Deployment Cost}
  \label{fig:cost_interlization_medium_cost}
  \resizebox{1\linewidth}{!}{\input{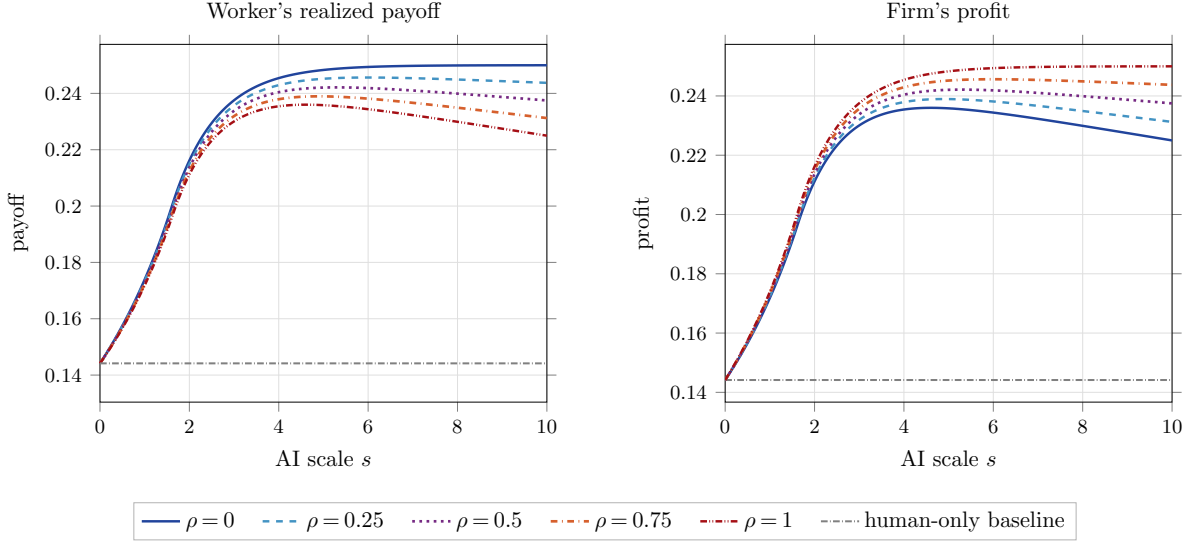}}
\end{figure}

\begin{figure}[ht]
  \centering
  \caption{Cost Internalization Under High AI Deployment Cost}
  \label{fig:cost_interlization_high_cost}
  \resizebox{1\linewidth}{!}{\input{fig/cost_internalization_r1_c02_ah1.tex}} 
\end{figure}

\smallskip
We first examine how the CI proportion $\rho$ affects worker payoff and firm profit under perfect collaboration, i.e., $\hat{\alpha} = \alpha$.
\cref{fig:cost_interlization_low_cost,fig:cost_interlization_medium_cost,fig:cost_interlization_high_cost} illustrate how shifting a larger share of AI deployment cost to workers affects both the worker's realized payoff and the firm's profit as AI scales up. Across all three cases, a common pattern emerges: a higher internalized cost share $\rho$ always reduces worker payoff, and when $\rho$ becomes sufficiently large the worker may optimally stop using the AI tool altogether, especially at higher AI scales. The effect on firm profit, however, depends on the cost environment. In the low-cost case (\cref{fig:cost_interlization_low_cost}), increasing $\rho$ generally benefits the firm, because the savings from shifting AI cost to the worker outweigh the relatively small reduction in worker participation. In the moderate-cost case (\cref{fig:cost_interlization_medium_cost}), this logic becomes weaker: although partial cost internalization can still improve firm profit, full internalization is no longer always optimal, particularly as AI scale increases, because excessive cost shifting discourages worker adoption and undermines the firm's gains from AI deployment. In the high-cost case (\cref{fig:cost_interlization_high_cost}), this trade-off becomes even sharper. The marginal benefit of further cost shifting is smaller, while the risk of pushing workers away from AI usage is much larger; as a result, firm profit is often higher when the firm bears part of the AI cost rather than setting $\rho=1$, even at intermediate AI scales. Managerially, these results suggest that cost internalization should not be treated as a one-size-fits-all policy. Aggressive cost shifting may be effective when AI is inexpensive, but when deployment cost is substantial, firms should absorb a meaningful share of the cost to preserve worker participation and sustain profit as AI scales.

\begin{figure}[ht]
  \centering
  \caption{Cost Internalization Under Worker Misperception}
  \label{fig:cost_interlization_misperception}
  \resizebox{1\linewidth}{!}{\input{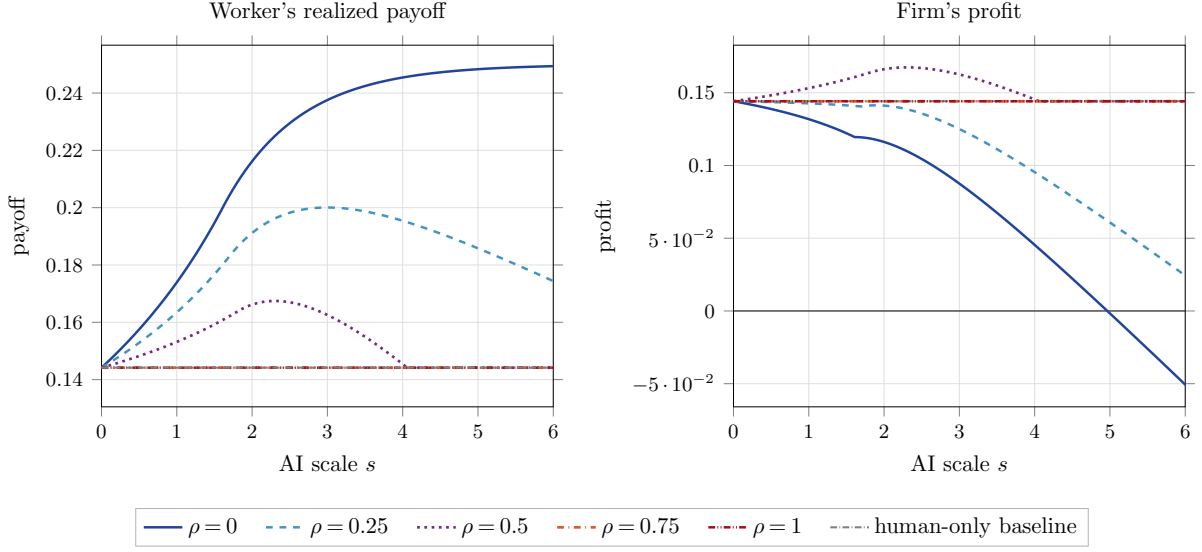}}
\end{figure}

\smallskip

Next, we examine the effect of cost sharing under worker misperception. \cref{fig:cost_interlization_misperception} illustrates how the worker's payoff and firm's profit change with different $\rho$ under fixed AI scale. 
We first focus on the left panel of \cref{fig:cost_interlization_misperception}, which shows the worker's realized payoff as the worker cost share $\rho$ increases. Under misperception, the curve has discontinuous points. These are the thresholds at which the worker stops using AI because the \emph{perceived} payoff from AI falls below the human-only payoff. Since the adoption decision is based on perceived payoff, while the realized payoff depends on the true AI capability, misperception creates a gap between the two and leads to these jumps. Notice that worker opts out of AI later when overperception is higher. Under perfect perception and under-perception, the worker's realized payoff decreases monotonically with $\rho$. Under over-perception, however, the realized payoff first decreases and then jumps upward at the discontinuity. This is because over-perception may cause the worker to rely too much on AI and cut human effort too aggressively, so the realized payoff can become very low, even lower than the human-only benchmark. Once the worker gives up AI, the payoff returns to the human-only level. This again echoes with the aforementioned results in \cref{prop:over_perception}.iii.

\smallskip

Turning to the right panel of \cref{fig:cost_interlization_misperception}, the firm's optimal cost-sharing policy also depends on worker misperception. Under over-perception, the firm's profit is maximized at a relatively high~$\rho$, meaning that the firm can shift a larger share of the AI cost to the worker. Intuitively, when workers already overestimate AI capability, they remain willing to adopt AI even when they bear more of the deployment cost, so stronger cost sharing allows the firm to save on AI expenditure without immediately losing adoption. At the same time, a higher $\rho$ also disciplines worker behavior by inducing them to retain more human effort, which partly mitigates the distortion caused by over-perception. By contrast, under under-perception, the profit-maximizing $\rho$ is lower, so the firm should bear a larger share of the AI cost itself. In this case, workers are already cautious about AI, and shifting too much cost to them further discourages adoption and reduces the firm's gains from AI deployment. Perfect perception lies between these two cases. Therefore, the optimal $\rho$ is not universal: when workers over-perceive AI capability, the firm can rely more on worker-side cost sharing, whereas when workers under-perceive AI, the firm should internalize more of the deployment cost.

\subsection{Perception Alignment}

\smallskip

Perception alignment refers to interventions that bring workers’ beliefs about AI capability closer to the true performance of the system. In our setting, it means reducing the gap between perceived capability $\hat{\alpha}$ and actual capability $\alpha$, so that workers choose effort  based on more accurate expectations. In practice, firms can implement perception alignment through calibrated training, transparent reporting of model performance and failure rates, benchmark-based evaluations on task-relevant use cases, and clear usage guidelines that communicate when AI should complement rather than replace human effort. 

\smallskip
\begin{figure}[ht]
  \centering
   \caption{Percentage improvement from perception alignment as AI scales,
    $(\text{aligned}-\text{misperceived})/\text{aligned}\times100\%$, for the worker's
    payoff (left) and the firm's profit (right), under symmetric misperception magnitudes
    $\hat\alpha-\alpha\in\{\pm0.2,\pm0.4,\pm0.6\}$. The dotted vertical line marks the true-$\alpha$ AGI threshold
    $s=\ln(1+t_0)/\alpha$. For the firm, under-perception (cool colors) can drive the
    improvement below zero, indicating that alignment would reduce profit. }
  \input{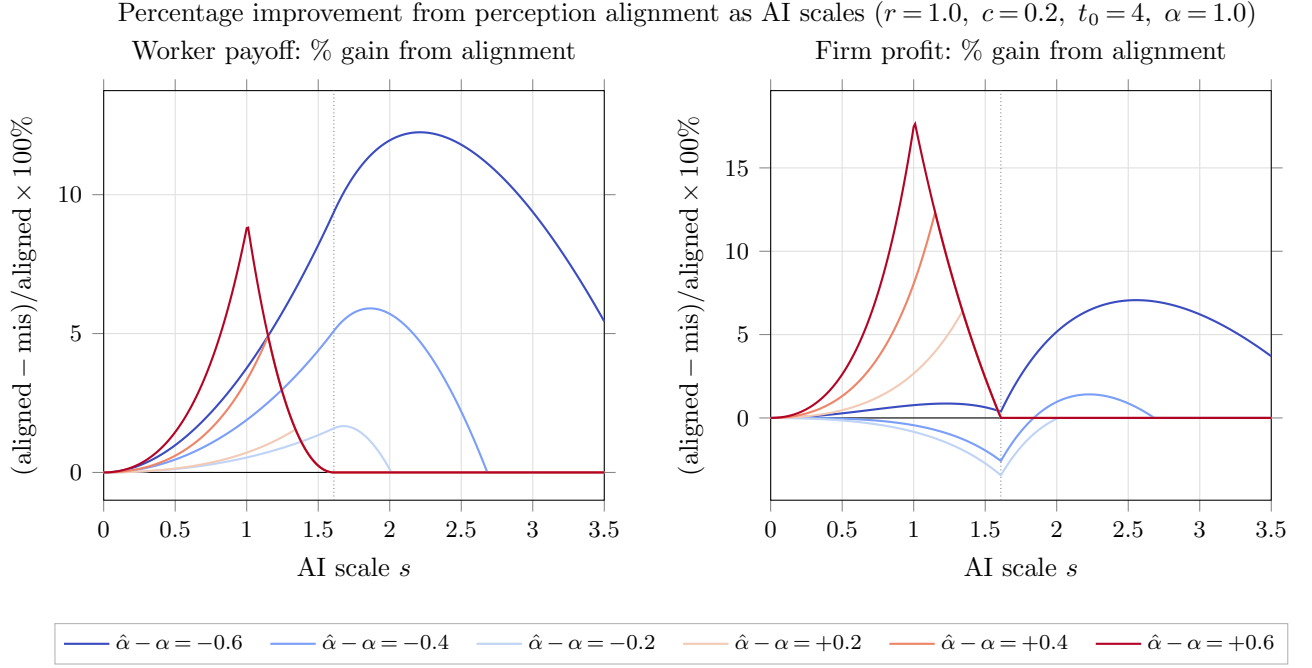}  
  \label{fig:align-pct}
\end{figure}

\Cref{fig:align-pct} shows the percentage improvement from perception alignment as AI scales, which reveals a sharp contrast between how perception alignment benefits the
\emph{worker} and the \emph{firm}. For the worker, alignment is unambiguously valuable under
both over- and under-perception: because the privately optimal effort $t^*(\alpha,s)$ maximizes
the worker's true payoff, correcting misperception in \emph{either} direction weakly raises the
realized reward $\hat R(\alpha,s)$ toward the benchmark $R^*(\alpha,s)$. The size of this gain is
governed by two factors: the \emph{magnitude} of misperception and the \emph{scale} of the AI.
Holding scale fixed, the percentage improvement rises monotonically with $|\hat\alpha-\alpha|$,
and it is concentrated in the HIL region of intermediate scale, peaking near the
worker's perceived AGI threshold $s=\ln(1+t_0)/\hat\alpha$ and vanishing once the AI is capable enough that the worker optimally exerts no effort. The worker therefore benefits most from
alignment when misperception is large and the AI operates at an intermediate scale; at very low or
very high scale, alignment is nearly inconsequential for the worker, and the direction of the
misperception is largely immaterial.

\smallskip

For the firm, by contrast, the value of perception alignment is markedly \emph{asymmetric} across
the two directions of misperception. Over-perception is unambiguously costly, and alignment
delivers reliable and increasing returns: the profit improvement is everywhere positive and grows
with the degree of over-perception. Under-perception, however, has an \emph{ambiguous} effect on profit. Since the worker does not internalize the AI cost, the firm always prefers more effort per project than the self-interested worker supplies, $t^{\mathrm{firm}}(\alpha,s)\ge t^*(\alpha,s)$; an under-perceiving worker over-invests effort and thereby unintentionally corrects this structural firm--worker misalignment that is created by their differing cost--reward structures. Consequently, aligning a mildly under-perceiving workforce  (e.g.,\ $\hat\alpha-\alpha\in\{-0.2,-0.4\}$) can actually \emph{reduce} firm profit over the relevant scale range---because alignment pulls the worker back from the firm-preferred effort level toward their own, lower optimum. Only when under-perception is severe enough that the worker overshoots even the firm's preferred effort (e.g.,\ $\hat\alpha-\alpha=-0.6$) does alignment again raise profit.

\smallskip

The central takeaway for managers is that the firm benefits far more \emph{certainly} from perception alignment when workers over-perceive AI capabilities. Investments in calibrating beliefs---training, transparent disclosure of true performance, and usage guidelines---pay off most clearly against over-reliance, where the returns are positive, monotone in the degree of misperception, and amplified at larger scale. Against under-perception, the case is conditional: a moderate degree of worker skepticism can be a feature rather than a bug, accidentally substituting for the costly incentive alignment the firm would otherwise have to engineer, so that aggressively ``de-biasing'' workers toward the true $\alpha$ may erode profit unless their underestimation is severe. 

\smallskip

Importantly, the consequences of over-perception are more pronounced at the firm level than at the level of the human-AI system alone. Because firms bear the cost of scaling while relying on human effort to sustain performance, miscalibrated beliefs can lead not only to lower realized rewards but also to inefficient over-deployment and amplified losses. In practice, however, organizations often pursue the opposite strategy: they scale up AI systems while simultaneously scaling down human involvement, increasing reliance on AI precisely when over-perception is most likely. Our analysis suggests that such practices may be more dangerous than they appear, as they can push firms deeper into the non-monotone regime where higher deployed scale may be associated with deteriorating outcomes.  
In this sense, investments in perception alignment can dominate investments in model scaling, particularly when workforce beliefs are prone to over-perception.

\section{Conclusion}
\label{sec:conclusion}
This paper examines when the empirical scaling law of AI translates into improved performance in human-AI collaboration. We show that the answer depends not only on advances in AI capability, but also on how humans respond to those advances. When workers accurately perceive AI capability, scaling enables them to reallocate effort efficiently, allowing the human-AI system to realize the technical gains from larger and more capable AI models. Human misperception, however, can fundamentally alter this relationship. Over-perception may generate a scaling paradox in which greater AI capability reduces overall system performance and amplifies firm-level profit losses, whereas under-perception generally preserves the benefits of scaling but slows their realization and can, in some cases, partially mitigate firm--worker incentive misalignment. We further show that firms can actively manage these distortions through operational policies such as cost internalization and perception alignment, whose effectiveness depends on the economics of AI deployment and the direction of human misperception. Taken together, our results suggest that AI scaling should be viewed not only as a technological challenge but also as an operational and behavioral one. Realizing the full value of increasingly capable AI systems requires firms to adapt human beliefs and incentive design to the capability of the AI system they deploy. 

\smallskip

As AI continues to advance, understanding the interaction between technological progress and human behavior through behavioral experiments will be increasingly important for ensuring that improvements in model capability translate into meaningful organizational performance. Another natural direction for future research is to endogenize the firm's scale decision by allowing it to choose \(s\) from a feasible set \(\mathcal{S}_{\tau}\) that expands over time as the AI technology frontier advances, thereby examining how worker misperception affects the timing and extent of AI adoption.

%

\medskip

\begin{APPENDICES}
\section{Proofs to Results in Section~\ref{sec:results-human}}

\subsection{Proof of Proposition~\ref{prop:optimal_t}}
Since \(C>0\) is a constant multiplicative factor, maximizing \(R(t;\alpha, s)\) is equivalent to maximizing
\[
g(t):=\frac{1-e^{-\alpha s-t}}{t+t_0}, \qquad t\ge 0\,.
\]
Differentiating yields
\[
g'(t)
=
\frac{e^{-\alpha s-t}(t+t_0)-\bigl(1-e^{-\alpha s-t}\bigr)}{(t+t_0)^2}
=
\frac{e^{-\alpha s-t}(t+t_0+1)-1}{(t+t_0)^2}\,.
\]
Define
\[
h(t):=e^{-\alpha s-t}(t+t_0+1)-1.
\]
Then
\[
g'(t)=\frac{h(t)}{(t+t_0)^2}.
\]
Moreover,
\[
h'(t)
=
-e^{-\alpha s-t}(t+t_0+1)+e^{-\alpha s-t}
=
-e^{-\alpha s-t}(t+t_0)<0
\]
for all \(t\ge 0\). Hence \(h\) is strictly decreasing on \([0,\infty)\). It follows that \(g'\) can vanish at most once, so \(g\) has at most one interior critical point; if such a point exists, it must be the unique global maximizer.

Now evaluate \(h\) at \(t=0\):
\[
h(0)=e^{-\alpha s}(t_0+1)-1.
\]

We distinguish two cases.

\medskip
\noindent\textbf{Case 1: \(\alpha s \ge \ln(1+t_0)\).}
Then \(h(0)\le 0\). Since \(h\) is strictly decreasing, we have \(h(t)<0\) for all \(t>0\). Therefore \(g'(t)<0\) for all \(t>0\), so \(g\) is strictly decreasing on \([0,\infty)\). Hence the unique maximizer is
\[
t^*(\alpha, s)=0.
\]

\medskip
\noindent\textbf{Case 2: \(\alpha s < \ln(1+t_0)\).}
Then \(h(0)>0\). Also,
\[
\lim_{t\to\infty} h(t) = -1.
\]
Since \(h\) is continuous and strictly decreasing, there exists a unique \(t^*>0\) such that \(h(t^*)=0\). This point is the unique global maximizer. The first-order condition is
\[
e^{-\alpha s-t^*}(t^*+t_0+1)=1.
\]

To solve explicitly, let
\[
y:=t^*+t_0+1.
\]
Then \(t^*=y-t_0-1\), and the first-order condition becomes
\[
e^{-\alpha s-(y-t_0-1)}\,y=1,
\]
or equivalently,
\[
(-y)e^{-y}=-e^{\alpha s-t_0-1}.
\]
Hence
\[
-y=W\!\left(-e^{\alpha s-t_0-1}\right),
\]
so
\[
t^*=-W\!\left(-e^{\alpha s-t_0-1}\right)-t_0-1.
\]

It remains to determine the correct real branch. Under \(\alpha s < \ln(1+t_0)\), we have
\[
-e^{\alpha s-t_0-1}\in(-1/e,0),
\]
so there are two real branches, \(W_0\) and \(W_{-1}\). Since \(t^*>0\), we need
\[
y=t^*+t_0+1>t_0+1\ge 1,
\]
hence
\[
-y\le -(t_0+1)\le -1.
\]
Therefore the relevant branch is \(W_{-1}\), because \(W_0(x)\in[-1,0)\) for \(x\in(-1/e,0)\), whereas \(W_{-1}(x)\in(-\infty,-1]\). Thus
\[
t^*(\alpha, s)=-W_{-1}\!\left(-e^{\alpha s-t_0-1}\right)-t_0-1.
\]

This proves the claim. \hfill $\blacksquare$

\subsection{Proof of Corollary~\ref{cor:monotonicity}}
\textbf{Monotonicity of $t^*$.}
In the interior (the non-zero effort region), implicit differentiation of the first-order condition $t + t_0 + 1 = e^{\alpha s + t}$ gives
\begin{equation}
\label{eq:optimal-t-partial-s}
\frac{dt^*}{ds} = -\frac{\alpha\, e^{\alpha s + t^*}}{e^{\alpha s + t^*} - 1} = - \alpha \frac{t^* + t_0 + 1}{t^* + t_0}< 0\,.
\end{equation}

At the boundary, $t^* = 0$ is constant. Hence $t^*$ is non-increasing in $s$ (and, by symmetry, in $\alpha$).

\noindent\textbf{Monotonicity of $R^*$.}
In the interior, the first-order condition gives $t^* + t_0 = e^{\alpha s + t^*} - 1$, so
\[
1 - e^{-\alpha s - t^*} = \frac{e^{\alpha s + t^*} - 1}{e^{\alpha s + t^*}} = \frac{t^* + t_0}{t^* + t_0 + 1}\,.
\]
Hence the optimal reward simplifies to
\[
R^* = \frac{t^* + t_0}{t^* + t_0 + 1} \cdot \frac{C}{t^* + t_0} = \frac{C}{t^* + t_0 + 1}\,.
\]
Since $t^*$ is strictly decreasing in $s$, $R^*$ is strictly increasing. At the boundary ($t^* = 0$), $R^* = (1 - e^{-\alpha s}) \cdot C / t_0$, which is also strictly increasing in $s$. Continuity at the transition $\alpha s = \ln(1 + t_0)$ is verified: both expressions yield $R^* = C/(1 + t_0)$. Hence $R^*$ is non-decreasing globally, and $R^*(\alpha, s) \geq R^*(0, 0)$ for all $\alpha, s \geq 0$. The claim regarding the number of completed projects follows from the same argument.
\hfill $\blacksquare$

\subsection{Proof of Proposition~\ref{prop:over_perception}}

\textbf{Part (i).} By \cref{prop:optimal_t}, $t^*(\alpha, s)$ is non-increasing in $\alpha$. Since $\hat{\alpha} > \alpha$, we have $\hat{t}(\hat{\alpha}, s) = t^*(\hat{\alpha}, s) \leq t^*(\alpha, s)$.

\noindent \textbf{Part (ii).} Since $R^*(\alpha, s) = \max_{t \geq 0} R(t; \alpha, s)$ and $\hat{t}$ is one particular feasible allocation, $\hat{R}(\alpha, s) = R(\hat{t}; \alpha, s) \leq R^*(\alpha, s)$.

\emph{Equality at $s = 0$}: when $s = 0$, both the perceived and true first-order conditions reduce to $t + t_0 + 1 = e^{t}$ (independent of $\alpha$ or $\hat{\alpha}$), so $\hat{t}(0) = t^*(0)$ and $\hat{R}(\alpha, 0) = R^*(\alpha, 0)$.

\emph{Equality for $s \geq \ln(1+t_0)/\alpha$}: by \cref{prop:optimal_t}, $\hat{\alpha} s > \alpha s \geq \ln(1+t_0)$ implies both $\hat{t} = 0$ and $t^* = 0$, giving $\hat{R}(\alpha, s) = R^*(\alpha, s) = (1 - e^{-\alpha s})\,C/t_0$.

\noindent  \textbf{Part (iii).} 
When $s=0$, $\hat{t}(\hat{\alpha}, s) = t^*(\alpha, s)$. Let $t_b^*$ denote the common optimum at $s = 0$, satisfying $t_b^* + t_0 + 1 = e^{t_b^*}$.

\emph{Initial increase}: differentiating $\hat{R}(\alpha, s) = (1 - e^{-\alpha s - \hat{t}}) \cdot C/(\hat{t} + t_0)$ and evaluating at $s = 0$, the chain rule gives
$$
\frac{d\hat{R}}{ds}(\alpha, 0) \;=\; \frac{C\alpha}{(t_b^*+t_0)(t_b^*+t_0+1)} \;>\; 0\,.
$$
Hence $\hat{R}(\alpha, s)$ is initially increasing at $s=0$.

\emph{Non-monotonicity for large $\hat{\alpha}/\alpha$}: at $s_b := \ln(1+t_0)/\hat{\alpha}$, the perceived allocation reaches the boundary $\hat{t}(s_b) = 0$, giving
$$
\hat{R}(\alpha, s_b) \;=\; \bigl(1 - (1+t_0)^{-\alpha/\hat{\alpha}}\bigr) \cdot \frac{C}{t_0}\,.
$$
As $\hat{\alpha}/\alpha \to \infty$, $(1+t_0)^{-\alpha/\hat{\alpha}} \to 1$, so $\hat{R}(\alpha, s_b) \to 0$, while $\hat{R}(\alpha, 0) = C/(t_b^* + t_0 + 1) > 0$ is fixed. By continuity of $\hat{R}$, there exists $\bar{\gamma} > 1$ such that $\hat{R}(\alpha, s_b) < \hat{R}(\alpha, 0)$ whenever $\hat{\alpha}/\alpha > \bar{\gamma}$, establishing non-monotonicity. Similarly by continuity, there exists $\tilde{s} < s_b$ such that $\hat{R}(\alpha, \tilde{s}) < \hat{R}(\alpha, 0)$. 

\emph{Recovery}: for $s \geq s_b$, $\hat{t} = 0$ and $\hat{R}(\alpha, s) = (1 - e^{-\alpha s})\,C/t_0$, which is strictly increasing toward $C/t_0$. For $s \geq \ln(1+t_0)/\alpha$, this coincides with $R^*(\alpha, s)$. 
\hfill $\blacksquare$

\subsection{Proof of \texorpdfstring{\cref{cor:over_perception_comparative_static}}{}}
Part (i) follows directly from \cref{cor:monotonicity}(i), which states that \(t^*(\alpha,s)\) is non-increasing in \(\alpha\). Since \(\alpha\le \hat{\alpha}_1<\hat{\alpha}_2\), we have
\[
t^*(\alpha,s)\ge t^*(\hat{\alpha}_1,s)\ge t^*(\hat{\alpha}_2,s),
\]
which is exactly the stated inequality.

For part (ii), fix \((\alpha,s)\) and consider the true reward as a function of human effort. By \cref{prop:optimal_t}, $R(t;\alpha,s)$ is quasi-concave in \(t\) and uniquely maximized at \(t^*(\alpha,s)\). Hence \(R(t;\alpha,s)\) is weakly increasing for all \(t\le t^*(\alpha,s)\). As a result, 
\[
R\!\left(t^*(\alpha,s);\alpha,s\right)
\ge
R\!\left(\hat{t}(\hat{\alpha}_1,s);\alpha,s\right)
\ge
R\!\left(\hat{t}(\hat{\alpha}_2,s);\alpha,s\right).
\]
This proves the claim. \hfill $\blacksquare$

\subsection{Proof of \texorpdfstring{\cref{prop:under_perception}}{}}

\textbf{Part (i).} Since $\hat{\alpha} < \alpha$, by the same monotonicity argument as Part (i) of \cref{prop:over_perception} (applied in the opposite direction), $\hat{t}(\hat{\alpha}, s) = t^*(\hat{\alpha}, s) \geq t^*(\alpha, s)$.

\noindent\textbf{Part (ii).} Differentiating $\hat{R}(\alpha,s) = R(\hat{t}(\hat{\alpha},s);\alpha,s)$ with respect to $s$:
$$
\frac{d}{ds}\hat{R}(\alpha,s) = \underbrace{\frac{\partial R}{\partial s}\bigg|_{\hat{t}}}_{\geq\,0} + \underbrace{\frac{\partial R}{\partial t}\bigg|_{\hat{t}}}_{\leq\,0} \cdot \underbrace{\frac{d\hat{t}}{ds}}_{\leq\,0}.
$$
The first term is non-negative since $\frac{\partial R}{\partial s} = \alpha\, e^{-\alpha s - t}\frac{C}{t+t_0} \geq 0$. The second term is non-negative because $\hat{t} \geq t^*$: the reward function is on the decreasing side of its maximum at $t^*$, so $\frac{\partial R}{\partial t}|_{\hat{t}} \leq 0$; combined with $\frac{d\hat{t}}{ds} \leq 0$ (from \cref{prop:optimal_t}), the product is non-negative. Hence $\frac{d}{ds}\hat{R}(\alpha,s) \geq 0$. Since $\hat{t} \geq t^*$ and $R$ is optimized at $t^*(\alpha,s)$, we have $\hat{R}(\alpha,s) = R(\hat{t};\alpha,s) \leq R(t^*;\alpha,s) = R^*(\alpha,s)$ for all $s > 0$. 
\hfill $\blacksquare$

\subsection{Proof of \texorpdfstring{\cref{cor:under_perception_comparative_static}}{}}
The proof follows the same arguments as the proof of \cref{cor:over_perception_comparative_static}.  \hfill $\blacksquare$

\section{Proofs to Results in \texorpdfstring{\cref{sec:firm}}{}}

\subsection{Proof of \texorpdfstring{\cref{lem:misalignment}}{}}

\textbf{Part~(i)}. For fixed \(s\), define $\Phi(t):=\bigl[r(1-e^{-\alpha s-t})-cs\bigr]\frac{C}{t+t_0}, \forall t\ge 0.$ Differentiating yields $\Phi'(t)=\frac{C}{(t+t_0)^2}\psi(t),$
where
\[
\psi(t):=re^{-\alpha s-t}(t+t_0+1)-r+cs\,.
\]
Moreover, \(\psi'(t)=-re^{-\alpha s-t}(t+t_0)<0\), so \(\psi\) is strictly decreasing on \(t\ge 0\). When \(\alpha s<\ln\!\left(\dfrac{t_0+1}{1-\frac{cs}{r}}\right)\), we have 
\[
\psi(0)=r\bigl(e^{-\alpha s}(t_0+1)-1\bigr)+cs>0.
\]
Also, since \(cs < r\), it holds that
\[
\lim_{t\to\infty}\psi(t)=-(r-cs)<0\,.
\]
Therefore, by continuity and strict monotonicity, there exists a unique \(t^{\mathrm{firm}}(\alpha,s)>0\) such that \(\psi(t^{\mathrm{firm}})=0\). This is equivalent to
\[
e^{-\alpha s-t^{\mathrm{firm}}}(t^{\mathrm{firm}}+t_0+1)=1-\frac{cs}{r}\,.
\]
Since \(\psi\) changes sign exactly once, \(\Phi'(t)>0\) for \(t<t^{\mathrm{firm}}\) and \(\Phi'(t)<0\) for \(t>t^{\mathrm{firm}}\). Thus \(\Phi\) is quasi-concave and has a unique interior maximizer \(t^{\mathrm{firm}}(\alpha,s)\). When \(\alpha s\geq\ln\!\left(\dfrac{t_0+1}{1-\frac{cs}{r}}\right)\), $\psi(t) \le 0$ for all $t \geq 0$. Thus, \(\Phi\) is non-increasing on $t \ge 0$ (and thus also quasi-concave) and has a unique maximizer $t^{\mathrm{firm}} = 0$.

\noindent\textbf{Part~(ii).} When $\alpha s \le \ln(t_0 + 1)$, by \cref{prop:optimal_t}, the human-optimal effort \(t^*(\alpha,s)\) satisfies \(e^{-\alpha s-t^*}(t^*+t_0+1)=1\) . By Part~(i), the firm's preferred effort satisfies \(e^{-\alpha s-t^{\mathrm{firm}}}(t^{\mathrm{firm}}+t_0+1)=1-\frac{cs}{r}<1\). Let \(h(t):=e^{-\alpha s-t}(t+t_0+1)\). Then \(h'(t)=-e^{-\alpha s-t}(t+t_0)<0\), so \(h\) is strictly decreasing. Since \(h(t^*)=1>1-\frac{cs}{r}=h(t^{\mathrm{firm}})\), it follows that \(t^{\mathrm{firm}}(\alpha,s)>t^*(\alpha,s)\). When $\alpha s > \ln(t_0 + 1)$, $t^* = 0$ and $t^{\mathrm{firm}} \ge 0$. The conclusion then follows. 
\hfill $\blacksquare$

\subsection{Proof of \texorpdfstring{\cref{prop:misalignment_overperception}}{}}

\textbf{Part (i).}
By \cref{lem:misalignment}, under the true parameter \(\alpha\),
\(t^{\mathrm{firm}}(\alpha,s)\ge t^*(\alpha,s).\) When \(\hat{\alpha}\ge \alpha\), by \cref{cor:monotonicity}(i), \(t^*(\cdot,s)\) is non-increasing in its first argument, so
\(t^*(\hat{\alpha},s)\le t^*(\alpha,s).\) Thus, it holds that 
\[
\Delta(\hat{\alpha}, s) = t^{\mathrm{firm}}(\alpha,s) - \hat{t}(\hat{\alpha},s)\,.
\]
Thus, $\frac{\partial \Delta(\hat{\alpha},s)}{\partial \hat{\alpha}} = -\frac{\partial \hat{t}(\hat{\alpha},s)}{\partial  \hat{\alpha}} \ge 0$.

\noindent\textbf{Part (ii).} By \cref{cor:monotonicity}(i), \(t^*(\hat\alpha,s)\) is weakly decreasing in \(s\). Next consider \(t^{\mathrm{firm}}(\alpha,s)\). On the interior region, Lemma~\ref{lem:misalignment} gives the characterization
\(e^{-\alpha s-t^{\mathrm{firm}}(\alpha,s)}
\bigl(t^{\mathrm{firm}}(\alpha,s)+t_0+1\bigr)=1-\frac{cs}{r}.\)
Implicit differentiation yields
\[
\frac{\partial t^{\mathrm{firm}}}{\partial s}
=
\frac{\frac{c}{r}(1+\alpha s)-\alpha}
{e^{-\alpha s-t^{\mathrm{firm}}(\alpha,s)}
\bigl(t^{\mathrm{firm}}(\alpha,s)+t_0\bigr)}.
\]
Since the denominator is positive on the interior region, the maintained condition
\[
\frac{c}{r}(1+\alpha s)\ge \alpha
\]
implies that \(t^{\mathrm{firm}}(\alpha,s)\) is weakly increasing on \(I\). Thus \(\Delta\) is weakly increasing locally at \(s \in I\).

\noindent\textbf{Part (iii).} When $s < \ln(t_0+1)/\hat{\alpha}$, we have $\frac{\partial \Delta(\hat{\alpha},s)}{\partial s} = \frac{\partial t^{\mathrm{firm}}(\alpha, s)}{\partial s} -\frac{\partial \hat{t}(\hat{\alpha}, s)}{\partial s}$. By \cref{cor:monotonicity}(i), it holds that 
\[\frac{\partial \hat{t}(\hat{\alpha}, s)}{\partial s} = -\frac{\hat{\alpha}\, e^{\hat{\alpha} s + \hat{t}}}{e^{\hat{\alpha} s + \hat{t}} - 1}\,.\]
Differentiating with respect to $\hat{\alpha}$ and using
$\frac{\partial \hat{t}}{\partial \hat{\alpha}}
=
- s\,\frac{e^{\hat{\alpha}s+\hat{t}}}
{e^{\hat{\alpha}s+\hat{t}}-1}$,
we obtain
\[
\frac{\partial^2 \hat{t}(\hat{\alpha},s)}
{\partial \hat{\alpha}\,\partial s}
=
-\frac{e^{\hat{\alpha}s+\hat{t}}}
{e^{\hat{\alpha}s+\hat{t}}-1}
-\hat{\alpha}s\,
\frac{e^{\hat{\alpha}s+\hat{t}}}
{\bigl(e^{\hat{\alpha}s+\hat{t}}-1\bigr)^3}<0\,.
\]
Hence $\frac{\partial \hat{t}(\hat{\alpha},s)}{\partial s}$
is strictly decreasing in $\hat{\alpha}$. 
\hfill $\blacksquare$

\subsection{Proof of \texorpdfstring{\cref{prop:profit_scaling_failure}}{}}
By definition of $\Pi(s; \alpha, \hat{\alpha})$, we have on the region \(\hat{\alpha} s<\ln(1+t_0)\): \[
\Pi'(s;\alpha,\hat{\alpha})
=
r\,\partial_s \hat{R}(\alpha,s)
-
c\Bigl(\hat{N}(\alpha,s)+s\,\partial_s\hat{N}(\alpha,s)\Bigr).
\]

By \cref{{cor:monotonicity}}.ii, $\partial_s\hat{N}(\alpha,s) \geq 0$. Since $\hat{N}(\alpha, s) > 0$, it holds that $\Pi'(s;\alpha,\hat{\alpha}) < r\,\partial_s \hat{R}(\alpha,s)$. 
\hfill $\blacksquare$

\subsection{Proof of \texorpdfstring{\cref{prop:misalignment_underperception}}{}}

Suppose \(\hat{\alpha}\le \alpha\) and \(\hat{\alpha}< \ln(t_0 + 1)/s\). Since \(t^*(\cdot,s)\) is strictly decreasing in its first argument on the interior region, there exists a unique threshold \(\hat{\alpha}^{\dagger}\) such that
\(t^*(\hat{\alpha}^{\dagger},s)=t^{\mathrm{firm}}(\alpha,s).\)
Using the human first-order condition,
\(e^{-\hat{\alpha}^{\dagger}s-t^{\mathrm{firm}}}(t^{\mathrm{firm}}+t_0+1)=1.\)
Combining this with the firm’s first-order condition from \cref{lem:misalignment},
\(e^{-\alpha s-t^{\mathrm{firm}}}(t^{\mathrm{firm}}+t_0+1)=1-\frac{cs}{r},\)
yields
\(e^{-\hat{\alpha}^{\dagger}s}
=
\frac{e^{-\alpha s}}{1-cs/r},\)
and hence
\(\hat{\alpha}^{\dagger}(\alpha,s)
=
\alpha+\frac{1}{s}\ln\!\left(1-\frac{cs}{r}\right).\)
Because \(t^*(\hat{\alpha},s)\) is strictly decreasing in \(\hat{\alpha}\), the two cases follow immediately.
\hfill $\blacksquare$

\end{APPENDICES}

{\footnotesize 
\bibliographystyle{informs2014} 

}

%

\end{document}